\RequirePackage{fix-cm}
\documentclass[twocolumn]{svjour3}          
\smartqed  
\usepackage{graphicx}
\usepackage{mathptmx}      
\usepackage{latexsym}
\usepackage{url}
\usepackage[pagebackref,breaklinks,colorlinks,citecolor=blue]{hyperref}
\usepackage{wrapfig}
\usepackage{graphicx}
\usepackage{amsmath}
\usepackage{amssymb}
\usepackage{booktabs}
\usepackage{dsfont}
\usepackage{algorithm}
\usepackage{makecell}
\usepackage{booktabs}
\usepackage{arydshln}
\usepackage{pifont}
\usepackage{array}
\usepackage{multirow}
\usepackage{tcolorbox}
\usepackage{makecell}
\usepackage{bm}
\usepackage{orcidlink}

\usepackage[authoryear]{natbib}

\usepackage{caption}

\newcommand{\vvb}[1]{\bm{#1}}

\newcommand{\pptxt}[1]{{\color{purple} #1}}
\newcommand{\sgntxt}[1]{{\color{green} $#1$}}
\newcommand{\ssbtxt}[1]{{\color{blue} $#1$}}

\newcommand{\ubtxt}[1]{{\textbf{\underline{#1}}}}

\newcommand{\eg}{\textit{e.g.}}
\newcommand{\ie}{\textit{i.e.,}}

\newcommand{\citeps}[1]{{{\small\citep{#1}}}}

\newcommand{\vsection}[1]{%
  \vspace{0pt}  
  \section{#1}  
  \vspace{0pt}  
}

\newcommand{\vsubsection}[1]{%
  \vspace{0pt}  
  \subsection{#1}  
  \vspace{0pt}  
}

\definecolor{DarkOrange}{RGB}{180,0,180}
\newcommand{\rrev}[1]{{#1}} 

\journalname{International Journal of Computer Vision}
\begin{document}

\title{HR-INR: Continuous Space-Time Video Super-Resolution via Event Camera}

\author{ \small
Yunfan~Lu$^1$ \orcidlink{0000-0002-7371-3189}
\and Yusheng~Wang$^2$ \and Zipeng~Wang$^3$ \and Pengteng~Li$^1$ \and Bin~Yang$^4$ \and Hui~Xiong*$^1$}

\institute{
Yunfan Lu \at
$^1$~AI Thrust, HKUST(GZ), China \\
\email{ylu066@connect.hkust-gz.edu.cn}
\and
Yusheng Wang \at
$^2$~University of Tokyo, Japan \\
\email{wang@robot.t.u-tokyo.ac.jp}
\and
Zipeng Wang \at
$^3$~HKUST, Hong Kong \\
\email{zwang253@cse.ust.hk}
\and
Pengteng Li \at
$^1$~AI Thrust, HKUST(GZ), China \\
\email{pli807@connect.hkust-gz.edu.cn}
\and
Bin Yang \at
$^4$~Aalborg University, Denmark \\
\email{byang@cs.aau.dk}
\and
\textbf{Hui Xiong *} \at
$^1$~AI Thrust, HKUST(GZ), China \\
\textit{Corresponding author} \\
\email{xionghui@ust.hk}
}

\date{Received: date / Accepted: date}

\maketitle
\begin{abstract}
Continuous space-time video super-resolution (C-STVSR) aims to simultaneously enhance video resolution and frame rate at an arbitrary scale.
Recently, implicit neural representation (INR) has been applied to video restoration, representing videos as implicit fields that can be decoded at an arbitrary scale.
However, existing INR-based C-STVSR methods typically rely on only two frames as input, leading to insufficient inter-frame motion information. Consequently, they struggle to capture fast, complex motion and long-term dependencies (spanning more than three frames), hindering their performance in dynamic scenes.
In this paper, we propose a novel C-STVSR framework, named HR-INR, which captures both \textbf{h}olistic dependencies and \textbf{r}egional motions based on INR.
It is assisted by an event camera -- a novel sensor renowned for its high temporal resolution and low latency.
To fully utilize the rich temporal information from events, we design a feature extraction consisting of (1) a regional event feature extractor -- taking events as inputs via the proposed event temporal pyramid representation to capture the regional nonlinear motion and (2) a holistic event-frame feature extractor for long-term dependence and continuity motion.
We then propose a novel INR-based decoder with spatiotemporal embeddings to capture long-term dependencies with a larger temporal perception field.
We validate the effectiveness and generalization of our method on four datasets (both simulated and real data), showing the superiority of our method.
The project page is available at {\url{https://github.com/yunfanLu/HR-INR}}.
\end{abstract}

\vsection{Introduction\label{sec:introduction}}

\begin{figure*}[t!]
\centering
\includegraphics[width=\textwidth]{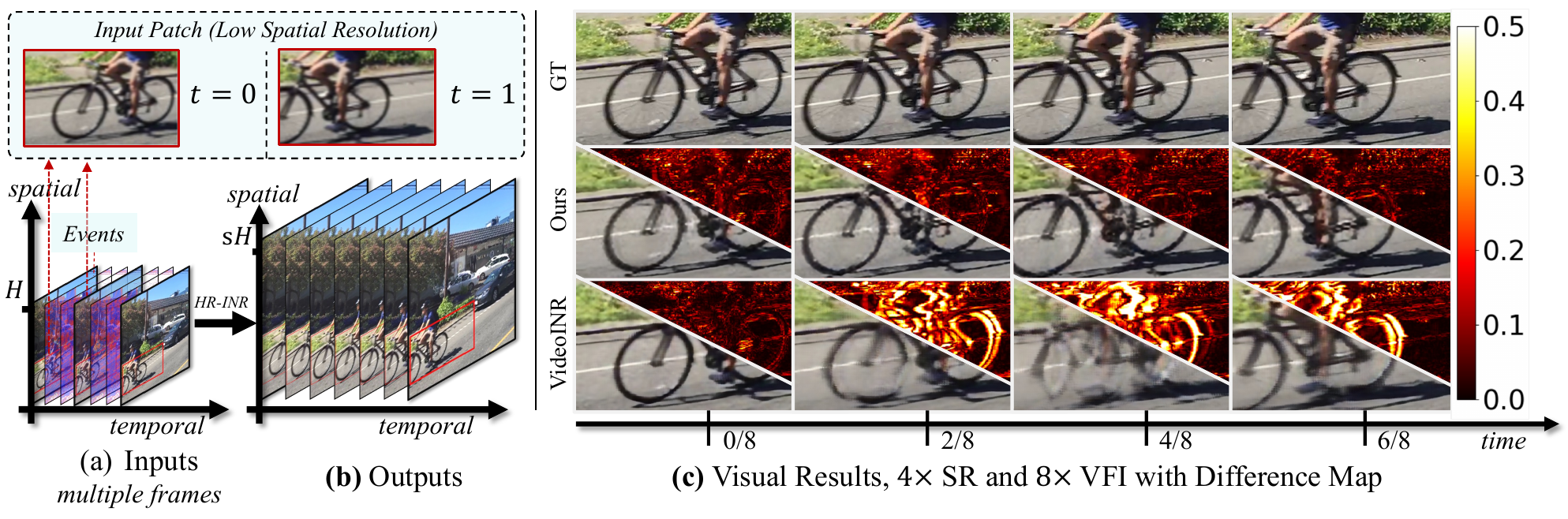}
\caption{
With event data as guidance, our method (HR-INR) takes in videos with low frame rates and resolution~\textbf{(a)} and produces continuous space-time videos with arbitrary frame rate and resolution~\textbf{(b)}.
Demonstrating effective modeling of local nonlinear motion, our method uniquely showcases this with the bicycle example in \textbf{(c)}, a feat unachievable by the prior method - VideoINR \citeps{chen2022videoinr}.
As shown in \textbf{(c)}, our method is able to recover the rotation of the bicycle wheels, which is unachievable by the prior method VideoINR \citeps{chen2022videoinr}.
}
\label{fig:cover-figure}
\centering
\end{figure*}

The real world's visual information, \eg, edge and object motion, is continuous, spanning both time and space dimensions.
However, the limited I/O bandwidth and sensor size of modern systems \citeps{delbracio2021mobile,parker2010algorithms} confines us to record videos at low frame rates and fixed resolutions.
This limitation has profound repercussions across various computer vision applications, \eg, encompassing immersive experiences in virtual reality \citeps{zhang2020and,lee2020experiencing}, traffic analysis in autonomous driving \citeps{zou2023object,zhao2019object}.
To address this limitation, recent research works, \eg, VideoINR \citeps{chen2022videoinr,chen2023motif}, have explored restoring videos by implicit neural representation (INR), with continuous resolutions and frame rates, referred to as Continuous Space-Time Video Super-Resolution (C-STVSR).

In practice, INR-based methods represent images or videos as neural fields that can be decoded at any resolution using a pointwise MLP decoder \citeps{cao2023ciaosr,chen2023cascaded}.
Once an image or video is encoded into this neural field, it effectively becomes a continuous function over space and time. Consequently, querying any previously unseen coordinate \((x,y,t)\) yields the corresponding RGB value, thus enabling arbitrary-scale spatial upsampling and arbitrary-frame-rate temporal interpolation.
One of the seminal INR approaches is LIIF \citeps{chen2021learning}, which is designed for arbitrary-scale image SR.
This line of research soon extended to the video domain.
In this context of C-STVSR, VideoINR \citeps{chen2022videoinr} employs a fixed STVSR model \citeps{xiang2020zooming} that extracts features from \ubtxt{two} consecutive video frames.
Then, it introduces a temporal INR to generate inverse backward warping optical flow \citeps{niklaus2020softmax} to warp features.
Lastly, it employs a spatial INR, similar to LIIF, to decode the RGB frame with arbitrary resolution.
Building upon this, MoTIF \citeps{chen2023motif} improves VideoINR by using forward motion estimation, reducing gaps and holes in the temporal INR, which are typically caused by the randomness and discontinuities associated with backward warping \citeps{park2021asymmetric}.
\textit{
However, these methods depend solely on \textbf{two} successive RGB frames, rendering the task of predicting inter-frame motions ill-posed, as shown in Fig.~\ref{fig:cover-figure} (c).
Consequently, it becomes challenging to accurately capture \textbf{highly dynamic motion} (\eg, regional high-speed or complex nonlinear movements) and to model \textbf{long-term dependencies} that extend across more than four frames.}

\begin{figure*}[t!]
\centering
\includegraphics[width=\linewidth]{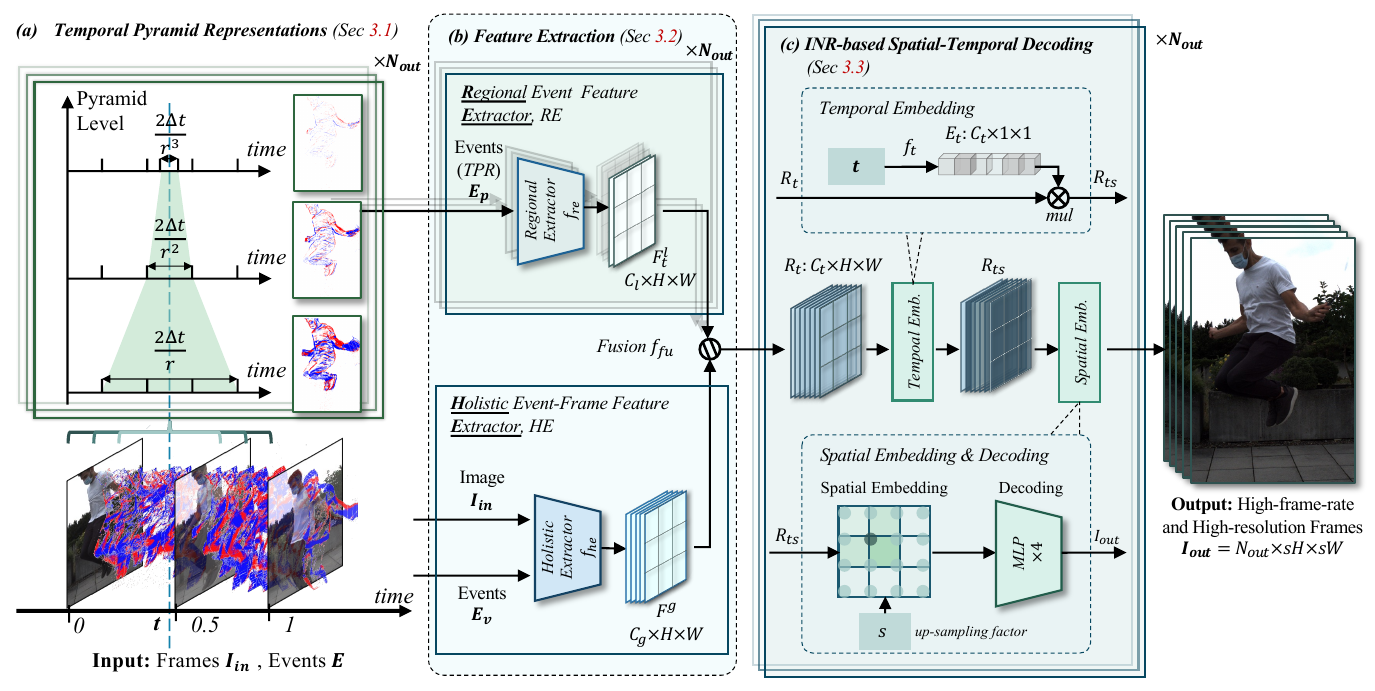}
\caption{
\textbf{Overview of our framework}.
The inputs are multi-frame images and their corresponding events.
The output is a video with enhanced frame rates and resolutions.
Firstly, events proximate to a particular time point are transformed into Temporal Pyramid Representations (TPR) to capture motion at a more granular temporal level \textbf{\textit{(a)}}.
Secondly, TPRs, the comprehensive set of multi-frames and events, are directed into the feature extraction part \textbf{\textit{(b)}}.
Within this part, the Regional Events Feature Extractor and the Holistic Events Feature Extractor process the input separately.
Lastly, the resulting features are then fused and inputted into an INR-based spatiotemporal decoding part \textbf{\textit{(c)}}.
Within this part, a temporal embedding is executed to capture features at a specific timestamp $t$, followed by spatial embedding with an up-sampling factor $s$ and decoding, culminating in the generation of frames at the desired resolution.}
\label{fig:over_all_framework}
\end{figure*}

\textbf{Motivation and Contributions:}
Event cameras are bioinspired sensors, known for their high temporal resolution and low latency ($<1 us$) \citeps{gallego2020event,zheng2023deep,lu2025events}.
Recent research has demonstrated the potential of events in guiding various video super-resolution (VSR) \citeps{lu2023learning,jing2021turning} and video frame interpolation (VFI) tasks \citeps{tulyakov2021time,tulyakov2022time,he2022timereplayer,chen2025repurposing,wang2025event,yan2025evstvsr,wei2025evenhancer}.
Specifically, \cite{jing2021turning} demonstrated that the high temporal resolution of events can be converted into high spatial resolution of frames.
Furthermore, \cite{lu2023learning} then extended this advantage to arbitrary-scale spatial upsampling.
 In the VFI, event streams provide motion information between frames.
\cite{tulyakov2021time,tulyakov2022time} proposed dual-stream methods to leverage the temporal information from events for interpolation at any given moment.
\cite{kim2023event} modeled finer motions through motion field estimation.
These methods have achieved significant progress in VFI and VSR tasks, and demonstrated how events can enhance both temporal information and spatial information.
However, utilizing events to facilitate joint video super-resolution and frame interpolation remains a challenging area that has not been thoroughly explored.
To end this, this paper introduces \textbf{HR-INR}, a novel INR-based method that leverages events for jointly guiding VSR and VFI through \textbf{h}olistic and \textbf{r}egional motion modeling.
It adeptly achieves C-STVSR by capturing regional, rapid motion and holistic, long-range motion dependencies, as shown in Fig.~\ref{fig:cover-figure}.
To capture the regional motion, we propose Temporal Pyramid Representation (TPR) to construct a time-series pyramid structure around the pivotal timestamp of events (Sec.~\ref{subsec:etpr}).
Different from the evenly divided timeline representations, like voxel grids \citeps{tulyakov2021time,tulyakov2022time,gallego2020event}, time surfaces \citeps{Sironi2018TimeSurface}, time moments \citeps{Han2021EvIntSR,lu2023learning} and symmetric cumulative \citeps{sun2022event}, TPR offers finer temporal granularity with less complexity and effectively captures rapid motion changes, as shown in Fig.~\ref{fig:over_all_framework}~(a).
Additionally, to estimate holistic, long-range motion (\textit{involving more than two frames}), our model can employ more frames (\eg, four frames) and their corresponding events.
This makes it possible to capture the extended-duration motion and improves the accuracy and continuity of motion modeling.

Accordingly, we design two specialized feature extractors: the \textbf{r}egional event feature \textbf{e}xtractor (RE) and the \textbf{h}olistic event-frame feature \textbf{e}xtractor (HE), see Sec.~\ref{subsec:encoding}.
Both extractors are grounded in the Swin-Transformer architecture \citeps{Liang2021SwinIR,Liu2022VSwin}, renowned for its efficacy and efficiency in video enhancement tasks.
The regional event feature extractor is a lightweight network designed specifically for extracting local information from our event TPR.
Meanwhile, {h}olistic event-frame feature {e}xtractor employs a more sophisticated approach, utilizing long-term and multi-scale fusion strategies to integrate both events and frames.
Consequently, our training and testing strategy requires HE to be invoked \textbf{only once} for multi-frame interpolation.
Subsequently, the extracted features from RE and HE are fused as the output of the feature extraction module.

After fusing the regional and holistic features, we propose a novel INR-based spatial-temporal decoding module (Sec.~\ref{subsec:inr}).
Our motivation is to reduce gaps and holes typically found in optical flow-based warping and multi-frame fusion, as show in Fig. ~\ref{fig:cover-figure} (c), which are also identified in previous research \citeps{chen2023motif,chen2022videoinr,he2022timereplayer,tulyakov2022time}.
To accomplish this, we propose an implicit temporal embedding designed to transform timestamps into focused attention vectors on long-term features.
This approach is crucial for effectively modeling long-term temporal dependencies and capturing regional dynamic motion and edges, ensuring attention is also given to long-distance dependencies.
Subsequently, we employ spatial embedding \citeps{chen2021learning} to achieve arbitrary up-sampling in the spatial dimension.

We conducted experiments on two simulated (Adobe240 \citep{Su_Delbracio_Wang_Sapiro_Heidrich_Wang_2017} and GoPro \citep{Nah_Kim_Lee_2017}) and two real-world datasets (BS-ERGB \citep{tulyakov2022time} and CED \citep{Scheerlinck_Rebecq_Stoffregen_Barnes_Mahony_Scaramuzza_2019}).
The results validate the superiority of our method and its excellent generalization capabilities on real-world datasets.
{Our approach is the \textbf{first} event-based method to achieve continuous space-time video super resolution, surpassing frame-based methods, as shown in Fig.~\ref{fig:cover-figure}.
It also \textbf{excels} in individual VSR and VFI metrics compared to previous event-based methods.}

\vsection{Related Works\label{sec:related-works}}
Since our research focuses on leveraging event data to guide the C-STVSR task, we categorize related works into five areas: (1) Frame-based VFI and VSR; (2) Frame-based Space-Time Video Super-Resolution; (3) INR for VFI and VSR; (4) Event-guided VFI and VSR; (5) Long-term Video Dependency Modeling. Sections (1) and (2) discuss the progress and limitations of frame-based methods, (3) highlights the role of INR in arbitrary scales, (4) covers event applications in VFI and VSR, and (5) emphasizes the importance of modeling long-term dependencies in videos.

\noindent\textbf{(1) Frame-based VFI and VSR}
VFI generates intermediate frames, while VSR improves video spatial resolution. Early frame-based works addressed these tasks separately. VSR focuses on modeling motion relations between multiple frames for feature alignment and resolution enhancement. Early methods \cite{bao2019memc,wang2019edvr} used motion compensation and deformable convolutions, while later works \citep{chan2021basicvsr,chan2022basicvsr++} explored longer temporal modeling. VFI, on the other hand, models inter-frame motion to generate intermediate frames, with early approaches relying on warping, kernels, or optical flow \citep{meyer2018phasenet,xu2019quadratic,dutta2021efficient,zhang2022optical}. Despite attempts to model nonlinear motion, these methods remain limited by the absence of inter-frame data. Recent work, such as RSST \citep{Liang2022RSST}, improves VFI by using multi-frame inputs.
In summary, both VSR and VFI rely on inter-frame motion modeling but are constrained by missing inter-frame motion information, making it challenging to capture complex nonlinear motion.

\noindent\textbf{(2) Frame-based Space-time Video Super-Resolution}
aims to enhance the resolution and frame rate of a video simultaneously \citeps{haris2020space,Kim_Oh_Kim_2020,xiang2020zooming,xu2021temporal}.
In comparison to two-stage solutions, where VFI~ \citeps{jiang2018super,xue2019video,niklaus2020softmax,niklaus2017video,cheng2020video}  and VSR~ \citeps{liu2018learning,yang2021real,yue2022real,wang2021unsupervised,tian2020tdan,isobe2020video} methods are applied sequentially, simultaneous space-time video super-resolution reduces cumulative errors and leverages the natural relations between VFI and VSR methods.
Zooming Slow-Mo \citeps{xiang2020zooming} uses temporal interpolation to generate missing frames and aligns temporal information using a deformable ConvLSTM network.
Similarly, TMNet \citeps{xu2021temporal} extracts short-term and long-term motion cues in videos by modulating convolution kernels.
\textit{However, these methods cannot simultaneously achieve spatiotemporal resolution across \textbf{arbitrary scales}. \ie C-STVSR.}

\noindent \textbf{(3) INR for VFI and VSR}
have achieved space-time video super-resolution with arbitrary resolutions \citeps{chen2022videoinr,chen2023motif} by learning videos implicit neural representations (INRs).
These methods primarily estimate optical flows from \ubtxt{two} consecutive frames to warp features into arbitrary space-time coordinates, which are then decoded using MLP layers.
\textit{However, by relying on just \textbf{two} consecutive frames, these methods cannot inherently model long-term motions (involving three or more frames) and fail to accurately capture local, inter-frame, non-linear motions due to missing inter-frame information.}

\noindent\textbf{(4) Event-guided VFI and VSR}
seek to boost performance by incorporating the biologically inspired event cameras~\citeps{zheng2023deep,lu2025rgb}.
Previous works have demonstrated the potential of event-guided VFI, which mainly focus on modeling non-linear motion with events~ \citeps{tulyakov2021time,tulyakov2022time,he2022timereplayer,song2023deblursr,chen2025repurposing,wang2025event,yan2025evstvsr,wei2025evenhancer}.
EFI \citeps{paikin2021efi} exclusively adopts the synthesis approach for intermediate frame generation.
TimeLens \citeps{tulyakov2021time} and TimeLens++ \citeps{tulyakov2022time} employ events to model nonlinear motion correlations, integrating both synthesis and warping-centric approaches.
Building on these advancements, CBM-Net \citeps{kim2023event}, introduces a motion field to handle complex movements.
E-CIR \citeps{song2022cir} and DeblurSR \citeps{song2023deblursr} use events to learn a function to map time to gray values, which can speed up the prediction time of each frame.
{However, while these VFI methods utilize events to capture inter-frame motion, they fail to establish long-term dependencies beyond two frames and support simultaneous VSR.}
The realm of event-guided video super-resolution has also been explored.
E-VSR \citeps{jing2021turning} highlights that high-frequency temporal information from events is beneficial to recovering high-frequency spatial information.
EG-VSR \citeps{lu2023learning} employs events to comprehend INR, allowing for video ups-sampling with arbitrary scale.
Contrasting these methods, we pioneer using events to enable concurrent VSR and VFI across arbitrary spatial-temporal scales, \ie C-STVSR.

\noindent\textbf{(5) Video Long-term Dependence Modeling} is a crucial aspect of VSR and VFI.
For instance, BasicVSR \citeps{chan2021basicvsr} and BasicVSR++\citeps{chan2021basicvsr++} enhance VSR performance by processing multi-frame inputs through an alignment module to model long-term motion correlations.
Similarly, in the VFI domain, many methods \citeps{suzuki2020residual,nah2019recurrent,zhang2020video} employ RNN or LSTM to model sequences of frames, capturing long-term dependencies effectively.
Furthermore, Zooming Slow-Mo \citeps{xiang2020zooming}, TMNet \citeps{xu2021temporal}, and RSST \citeps{Liang2022RSST} leverage multi-frame inputs in the joint task of VSR and VFI, showcasing the importance of integrating multiple frames for improved modeling of video dynamics.
\textit{However, current C-STVSR methods \citeps{chen2022videoinr,chen2023motif}, and event-based VFI methods \citeps{tulyakov2021time,he2022timereplayer,tulyakov2022time,kim2023event}, primarily rely on estimating optical flow between two consecutive frames, which is then used to warp these two frames to generate the intermediate frames.
Therefore, they are challenging to handle multi-frames as inputs, inherently undermining their capability to model long-term dependencies.}

\vsection{Proposed Framework\label{sec:methods}}

\noindent\textbf{Preliminary:}
Events are discrete points that capture the positive and negative changes in pixel brightness.
Their generation depends on brightness alterations within the logarithmic domain.
Specifically, an event point $e=(x,y,t,p)$ is triggered and logged upon meeting certain criteria.
Suppose $L(x,y,t)$ represents the brightness at point $(x,y)$ at any given time $t$.
The event is recorded if the absolute difference $\Delta L = log\left(L(x,y,t)\right) - L\left(x,y,t - \Delta t\right)$ surpasses a predetermined threshold $C$, as formulated as Eq.~\ref{eq:event_generation}.
\begin{equation}
    \begin{aligned}
        p = \left\{
            \begin{aligned}
                +1, &\Delta L > C \\
                -1, &\Delta L < -C
            \end{aligned}
        \right.
    \end{aligned}
    \label{eq:event_generation}
\end{equation}

Utilizing Eq.~\ref{eq:event_generation}, the processing pipeline for a specified pixel at coordinates $(x, y)$ at any given time $t$ and $t'$ can be delineated by Eq.~\ref{eq:frame_relation_with_event}.

\begin{equation}
   \begin{aligned}
        L(x,y,t) = I(x,y,t') \times \exp({C \int_{t'}^{t} p~ dt})
    \end{aligned}
    \label{eq:frame_relation_with_event}
\end{equation}

Utilizing Eq.~\ref{eq:frame_relation_with_event} along with corresponding events, intensity frames at a given time can be used to compute frames at alternate times, facilitating video frame interpolation. However, events are often noisy, and directly applying Eq.~\ref{eq:frame_relation_with_event} may not yield optimal results \citeps{pan2019bringing}.
Therefore, employing neural networks for event-based frame interpolation has garnered significant attention \citep{paikin2021efi,tulyakov2021time,zhang2022unifying,kim2023event}. Additionally, the high temporal resolution of events helps in converting to high spatial resolution, as demonstrated by previous studies \citeps{lu2023learning,jing2021turning}.
In conclusion, event signals play a crucial role in frame interpolation and video super-resolution tasks, making them a natural guide for continuous space-time video super-resolution.

\noindent\textbf{Overview of HR-INR:} Our proposed HR-INR framework is depicted in Fig.~\ref{fig:over_all_framework}.
The inputs of this framework are RGB frames $\vvb{I}_{in}=\{I_{in}\}_i^{N_{in}}\in R^{N_{in}\times H\times W\times 3}$ and associated events $\vvb{E}$.
$H$ and $W$ denote the spatial resolution of frames and events.
$3$ means three channels of RGB.
$N_{in}$ denotes the input number of frames.
Furthermore, the framework outputs a video with an arbitrary frame rate and spatial resolution.
In particular, we consider the output video as $\vvb{I}_{out}=\{I_{out}\}_i^{N_{out}}$, consists of $N_{out}$ frames, each with a resolution of $(s\times H)\times (s\times W)$, where $s$ represents the up-sampling scale greater than $1$.
For the output $N_{out}$ frames, we denote the time corresponding to each frame as $\vvb{T}=\{t\}_i^{N_{out}}$.
For convenience, we also record the up-sampling scale $s$ and the time $\vvb{T}$ as a part of inputs.
Therefore, the mapping function $f_{hr}(.)$ of C-STVSR can be described by Eq.~\ref{eq:hr-inr}.

\begin{equation}
    \vvb{I}_{out} = f_{hr}\left(\vvb{I}_{in},\vvb{E}, s, \vvb{T}\right) \label{eq:hr-inr}
\end{equation}

Our framework comprises three main components:
First, Sec.~\ref{subsec:etpr} presents the event temporal pyramid representation (TPR), capturing regional dynamic motion and edges.
Second, Sec.~\ref{subsec:encoding} elaborates on the feature extraction process using regional and holistic feature extractors.
Third, Sec.~\ref{subsec:inr} describes the INR-based spatiotemporal decoding.

\begin{figure}
\centering
\includegraphics[width=\linewidth]{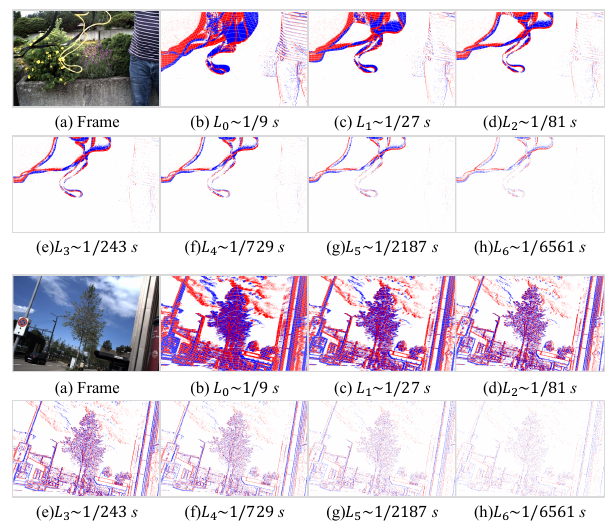}
\caption{Visualization of Event TPR across different time resolutions. The TPR is divided into 7 layers ($L=7$), with each layer having a time resolution that is $1/3$ of the previous layer ($r = 3$). The time resolution for the first layer ($L_0$) is approximately $2\times \Delta t = 1/9 s$, and the time resolution for the seventh layer ($L_6$) is approximately $1/6561 s$. Each layer visualizes the corresponding event data at a specific time resolution, demonstrating the effect of varying temporal resolution on event-based data.}
\label{fig:21-TPR}
\end{figure}

\vsubsection{Temporal Pyramid Representation\label{subsec:etpr}}
Firstly, we represent the event stream $\vvb{E}$ into a frame-like form that the network can process.
To capture holistic motion, we partition all events during $[0,1]$ of the timeline into $M$ equal intervals using a voxel grid \citeps{tulyakov2021time,gallego2020event}, denotes as $\vvb{E}_v$ with dimensions $M \times H \times W$.
In practice, event representation methods like voxel grid \citeps{tulyakov2021time,gallego2020event} and its extended structure, symmetric cumulative event representation \citeps{sun2022event},  achieve time granularity by uniformly dividing time into intervals with resolutions of $1/M$.

However, for C-STVSR, capturing intricate motion and edges requires a finer granularity.
Merely increasing $M$ to enhance detail sharply raises computational costs; for instance, capturing $1/1000$ second intervals within a second necessitates expanding $M$ to $1000$, which is computationally impractical.
To address this, we introduce Temporal Pyramid Representation (TPR), leveraging the high temporal resolution of events while reducing representation complexity.

\noindent
\textbf{Our idea:} The core of TPR is constructing a temporal pyramid where each successive layer's duration is $1/r, (r > 1)$ of the preceding one, leading to exponentially finer time granularity with additional layers.
For instance, as illustrated in Fig.~\ref{fig:over_all_framework}~(a), around any given time $t$, we define a surrounding time window $\Delta t$ and select an attenuation factor $r$.
At the pyramid's $L$-th level, the events are within the time span of $[t - \Delta t / r^{L}, t + \Delta t / r^{L}]$.
Each layer is further segmented into $M_p$ intervals, represented using a voxel grid.
Accordingly, for an $L$-th layer, each layer contains $M_p$ moments within the TPR, and its finest time granularity, denoted by $\delta_t$, is as delineated in the Eq.~\ref{eq:time_granularity}.
Therefore, for any time $t$, we construct the corresponding TPR $E_p$ with shape $L \times M_p \times H \times W$.
A simplified example is shown in Fig. \ref{fig:21-TPR}.
We record the TPRs at all target timestamp as $\vvb{E_p}=\{E_p\}_i^{N_{out}}$.

\begin{equation}
    \delta_t = \frac{2\times \Delta t}{M_p \times r^{L}} \label{eq:time_granularity}
\end{equation}

\noindent
\textbf{Discussion:}
The time granularity $\delta_t$ of TPR exponentially improves with the increase in layers - $L$.
For an attenuation factor of $r=3$ and a goal to detect motions as brief as $1/1000$ of a second within a $1s$ window ($2 \times \Delta t = 1$), we require only $7$ TPR layers with each layer divided into $2$ intervals ($M_p = 2$).
Consequently, a TPR with dimensions $7\times 2\times H\times W$ suffices to discern motion down to $1/1000 s$.
Based on the above representations, we obtained $\vvb{E}_v$, encapsulating holistic motion, and $\vvb{E}_p$, which focuses on regional edges and motion.

\vsubsection{Holistic-Regional Feature Extraction\label{subsec:encoding}}
This module aims to extract features from regional TPRs $\vvb{E_p}$, and the frame $\vvb{I_{in}}$ and the holistic events $\vvb{E_v}$ for INR-based spatial-temporal decoding.
Accordingly, we introduce:
\textbf{(1)} the regional event feature extractor $f_{re}$, Eq.~\ref{eq:feature_extraction} (a), for dynamic motion and edges detail capture.
\textbf{(2)} the holistic event-frame feature extractor $f_{he}$, Eq.~\ref{eq:feature_extraction} (b), for long-term motion dependencies modeling across time and space.

\begin{figure}[t!]
\centering
\includegraphics[width=\linewidth]{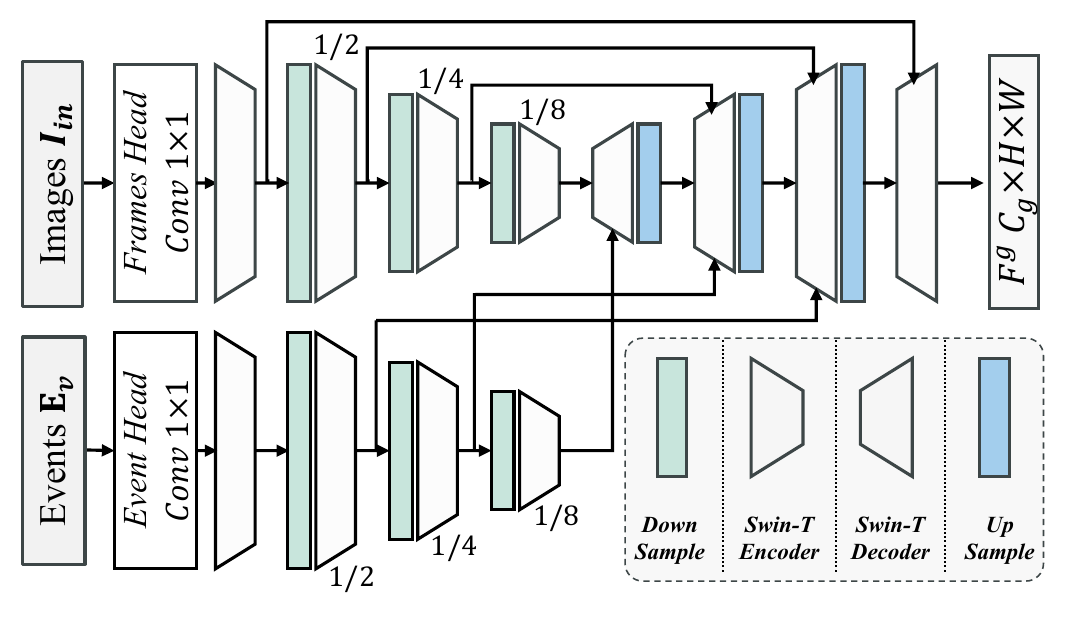}
\caption{
Holistic event-frame feature extractor. The down-sample module will halve the resolution. The up-sample module will double the resolution. The encoder and decoder have the same structure as Swin-Transformer \citeps{liu2022video,liu2021swin,geng2022rstt}.}
\label{fig:Encoder}
\centering
\end{figure}
\begin{equation}
\begin{split}
    F_t^l =&~ f_{re}\left(\vvb{E_p}\right) & (a)\\
    F^g =&~ f_{he}\left(\vvb{I_{in}}, \vvb{E_v}\right) & ~~~(b)\\
    R_t =&~ f_{fu}\left(F^g, F_t^l\right) & (c)\\
    \label{eq:feature_extraction}
\end{split}
\end{equation}

\begin{figure*}[t!]
\centering
\includegraphics[width=0.9\linewidth]{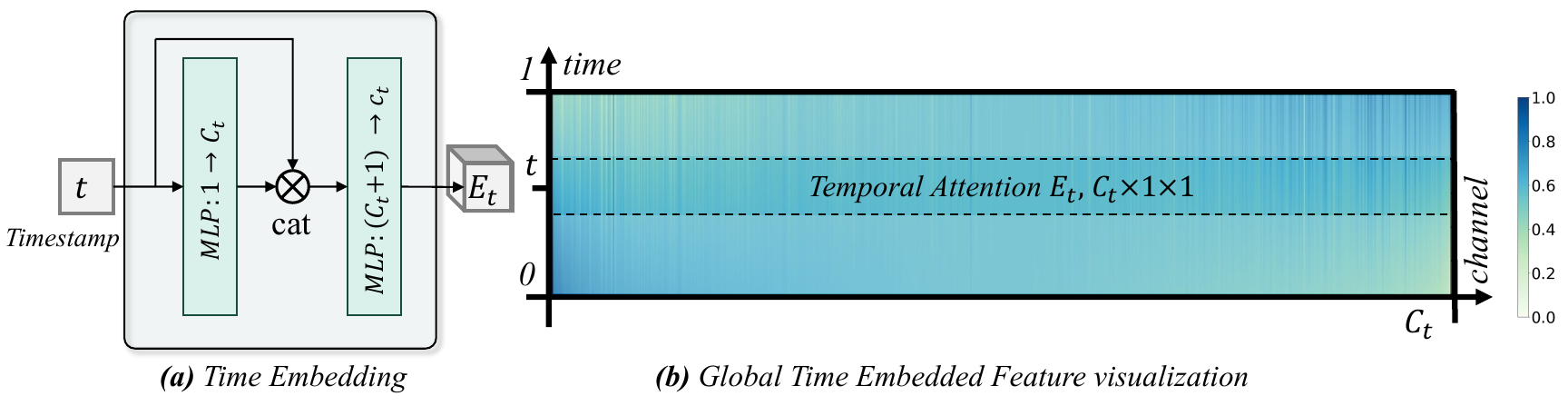}
\caption{
Temporal embedding. Given the input time $t \in [0,1]$, the output is the temporal attention $E_t$ derived from a two-layer MLP. \textbf{(b)} presents a visualization of the trained $E_t$ during $[0,1]$ on real-world dataset \citeps{tulyakov2022time}.}
\label{fig:8-TimeEmbeddingFeatureVisualizate}
\label{fig:8-TFV}
\centering
\end{figure*}
\noindent \textbf{(1) Regional Event Feature Extraction:}
The regional event feature extractor \( f_{re} \) takes as input the Temporal Pyramid Representation (TPR) \( \vvb{E_p} \in \mathbb{R}^{L \times M_p \times H \times W} \) at each timestamp \( t \), where \( L \) denotes the number of TPR layers and \( M_p \) represents the number of moments within each layer. Since \( f_{re} \) is invoked \( N_{out} \) times, its design must balance computational efficiency with the capacity to model inter-level relationships effectively, ensuring the accurate capture of regional motion and edge details.
The feature extraction process is divided into two stages: preprocessing and the main feature extraction via Swin Transformer Encoder Blocks (STEB) \citeps{liu2022video,liu2021swin,geng2022rstt}. In the preprocessing stage, a \( 1 \times 1 \) convolutional layer is applied to increase the feature dimensions to \( C_r \), resulting in a shape of \( L \times C_r \times H \times W \). This step enhances the feature's capacity to capture intricate details from the event stream.
In the subsequent feature extraction stage, four STEBs are employed to model the relationships between different pyramid levels. STEB utilizes a multi-head self-attention mechanism, which enables the model to effectively capture long-range dependencies and convey edge information across different levels of the pyramid. This mechanism improves the model's ability to capture short-term motion while maintaining computational efficiency through parameter optimization. The input and output dimensions of STEB remain consistent throughout the process, preserving the feature shape \( L \times C_r \times H \times W \).
Thus, the \( f_{re} \) module incorporates both preprocessing and feature extraction steps to ensure efficient extraction of regional motion features while maintaining high computational efficiency.

\noindent\textit{Swin Transformer Encoder Blocks (STEB):}
The input and output shapes of STEB are represented as $L\times C\times H\times W$ for clarity.
Initially, for a window size of $M\times M$, specifically $4\times 4$ in our implementation, the input is partitioned into disjoint windows of $(M \times M) \times L \times (H/M) \times (W/M) \times C$ dimensions.
Then, each window is compressed to form a feature map of shape $(M\times M)\times (L\times H\times W / M^2) \times C$.
Following this, Layer Normalization~\citeps{ba2016layer} and window-based multi-head self-attention~\citeps{liu2021swin,geng2022rstt} are computed for each window, succeeded by further transformation via another Layer Normalization and a Multi-Layer Perception.
Shifted window-based multi-head self-attention~\citeps{liu2021swin,geng2022rstt} is then employed to establish cross-window connections.
After applying one STEB structure, a second STEB is introduced with an identical configuration, except the input feature window is offset by $(M/2)\times (M/2)$.
In total, four STEBs are utilized for comprehensive feature extraction in the Regional Event Feature Extraction.

\noindent\textbf{(2) Holistic Event-Frame Feature Extraction:}
The inputs of $f_{he}$ is $\vvb{I_{in}}$ and $\vvb{E_v}$, as shown in Fig.~\ref{fig:Encoder}.
First, a convolutional layer processes both frames and events to increase dimensions.
Motivated by the events feature manifest between successive frames to provide inter-frame motion information.
We construct $N_{in}-1$ event segments so that each segment complements the inter-frame motion between two neighboring RGB frames.
We then incorporate the STEB to facilitate interactions across varied levels and spatial domains.
To minimize computational overhead and expand the receptive field, we integrated a down-sampling module between STEBs, forming a multi-scale encoder.
Each down-sampling iteration halves the resolution while maintaining the channel dimensions.
After three iterations, the feature resolution reduces to $1/8$ of its initial size, enlarging the receptive field.
We then employ a Swin Transformer Decoder Block \citeps{liu2022video,liu2021swin,geng2022rstt} to fuse features at matching resolutions.
Each of the first three STEBs is followed by an upsampling process, which doubles the resolution while maintaining the channel count.
Ultimately, this process outputs the feature $F_g$.

\noindent\textbf{Outputs of Feature Extraction:} To output $N_{out}$ frames, the holistic event-frame feature extractor $f_{he}$ is called once, capturing the comprehensive feature $F^g$.
Subsequently, for each time $t$, $f_{re}$ extracts regional features $F_t^l$ from each TPR $E_p \in \vvb{E_p}$.
For each regional feature $F_t^l$, we use addition and \textit{Conv}$1\times 1$ operation as fusion function $f_{fu}$ to fuse with holistic feature $F^g$ to obtain the output $R_t$ as Eq. \ref{eq:feature_extraction} (c).
For each time $t$, the whole process can be described by Eq.~\ref{eq:feature_extraction}.

\begin{figure*}[t!]
\centering
\includegraphics[width=\linewidth]{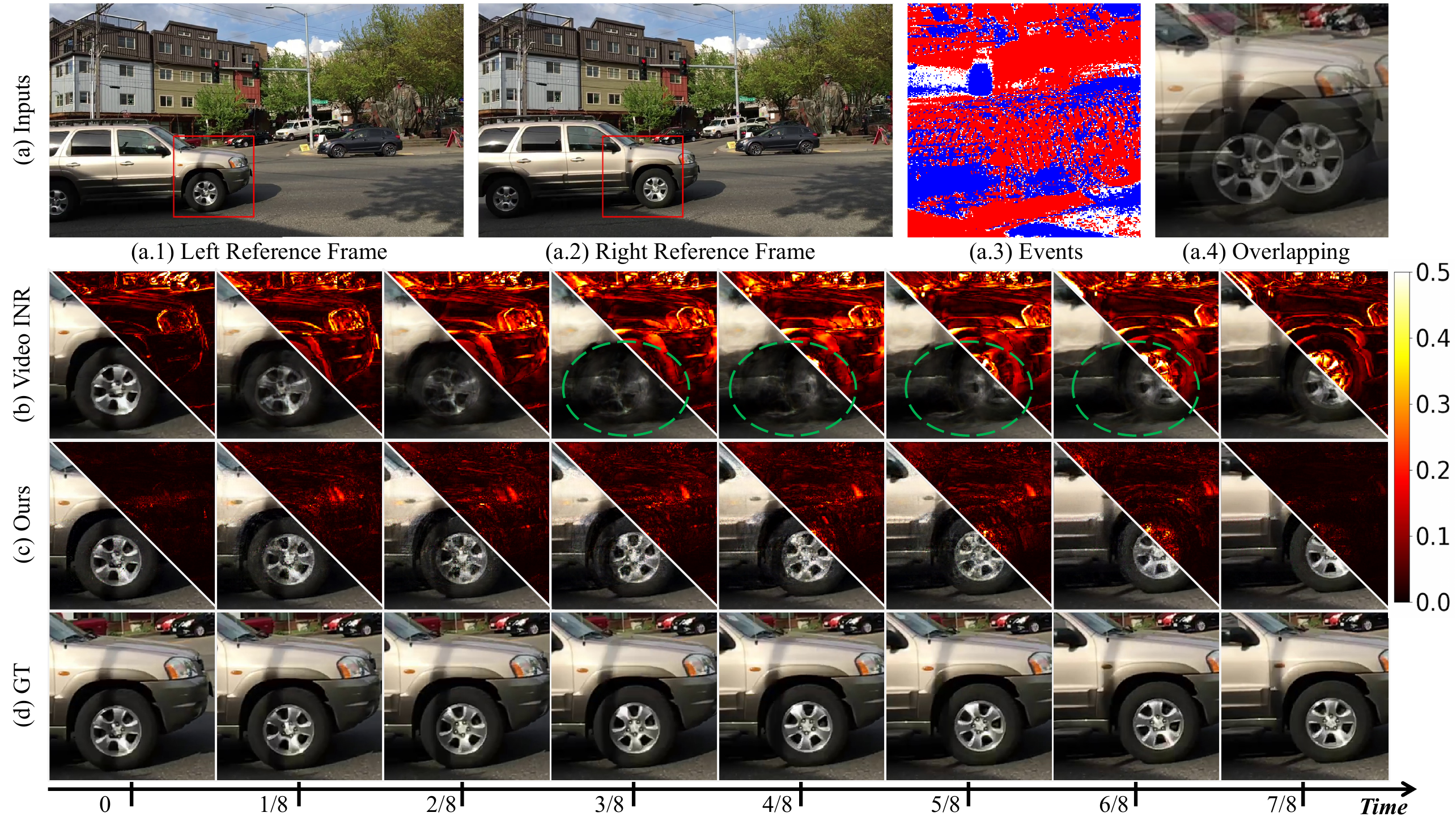}
\caption{7-\textit{skip} frame interpolation visualization results in Adobe240 dataset \citeps{Su_Delbracio_Wang_Sapiro_Heidrich_Wang_2017}. Our method \textbf{(c)} more effectively captures the \textbf{rotating wheels} compared to the VideoINR \textbf{(b)}, which tends to show noticeable holes and gaps. {\color{green} Green circles} highlight obvious holes and gaps.}
\label{fig:7-ADB-VFI-Vis-Release}
\centering
\end{figure*}

\vsubsection{INR-based Spatial-Temporal Decoding\label{subsec:inr}}
In this section, we employ INR-based spatial-temporal decoding to effectively retrieve RGB frames at any desired time and resolution. To achieve this, we leverage a temporal INR to generate features at any timestamp and a spatial INR to upscale the features to any spatial resolution.

\noindent\textbf{Temporal Embedding:}
We utilize learned temporal embedding as attention vectors to aggregate the fused feature in the channel dimension in a time-specific manner.
At a given timestamp $t$, we first use a two-layer MLP to increase its dimension to $C_t$, resulting in a temporal attention vector (as illustrated in Fig.~\ref{fig:8-TimeEmbeddingFeatureVisualizate}~(a)).
The visualization of the learned temporal attention is depicted in Fig.~\ref{fig:8-TFV} (b), exhibiting variations across both time and channel dimensions.
This attention vector is then multiplied directly with $R_t$ to generate the temporal embedded feature $R_{ts}$.
This temporal INR allows for a larger temporal perception field without the need for estimating optical flows.
Next, a $1\times 1$ convolution is applied to compress $R_{ts}$ to $C_{ts}$ dimensions, reducing the complexity of the spatial embedding and decoding.

\noindent\textbf{Spatial Embedding and Decoding:}
To upscale the temporal embedded features to any desired spatial scale, we utilize a similar approach to previous works \citeps{chen2022videoinr,chen2023motif,chen2021learning}.
We query the four nearest neighbors in the temporal embedded feature for each spatial coordinate and concatenate these with their distances for spatial embedding.
A four-layer MLP decoder computes the RGB values, which are then aggregated through area-weighted interpolation for arbitrary-scale super-resolution.
Similar to \citeps{lu2023learning}, we use the \textit{Charbonnier loss} \citeps{lai2018fast} as the fundamental loss function for training.

\vsection{Experiments\label{sec:experiments}}
In this section, we first introduce the datasets, network training and testing details, as well as the evaluation methods in Sec. \ref{sec:exp:setting}.
Then, we conduct comparative experiments, which include both joint C-STVSR comparisons and individual comparisons for VFI and VSR tasks in Sec. \ref{sec:comparisino}.
Furthermore, in the following analysis and ablation studies, we explore the performance and characteristics of various components within our framework, as discussed in Sec. \ref{sec:analytical_experiments}.

\subsection{Datasets and Implementation Setting\label{sec:exp:setting}}

\begin{table*}[t!]
\centering
\caption{Quantitative metrics (PSNR/SSIM) with 7-\textit{skip} VFI and $4\times$ VSR. \textit{Center} \textit{Average} remain consistent with previous work~\citeps{chen2022videoinr,chen2023motif}.\label{tab:comp-gopro-adobe-f-stvsr}}
\resizebox{1\linewidth}{!}{
\setlength{\tabcolsep}{0.022\linewidth}{
\begin{tabular}{lll|ccccc}
\toprule
VFI        & VSR      & Params~($M$)   & GoPro-\textit{Center}    & GoPro-\textit{Average}     & Adobe-\textit{Center}     & Adobe-\textit{Average}  \\
\midrule
\hline
                                    & Bicubic                           & 19.8                      & 27.04/0.7937  & 26.06/0.7720    & 26.09/0.7435    & 25.29/0.7279 \\
\makecell[l]{Su-SloMo}    & 
EDVR           & 19.8+20.7                 & 28.24/0.8322  & 26.30/0.7960    & 27.25/0.7972    & 25.95/0.7682 \\
                                    & BasicVSR   & 19.8+6.3                  & 28.23/0.8308  & 26.36/0.7977    & 27.28/0.7961    & 25.94/0.7679 \\
\hdashline
                                    & Bicubic                           & 29.2                      & 26.50/0.7791  & 25.41/0.7554    & 25.57/0.7324    & 24.72/0.7114 \\
\makecell[l]{QVI} 
& EDVR           & 29.2+20.7                 & 27.43/0.8081  & 25.55/0.7739    & 26.40/0.7692    & 25.09/0.7406 \\
                                    & BasicVSR   & 29.2+6.3                 & 27.44/0.8070  & 26.27/0.7955    & 26.43/0.7682    & 25.20/0.7421 \\
\hdashline
                                    & Bicubic                           & 24.0                      & 26.92/0.7911  & 26.11/0.7740    & 26.01/0.7461    & 25.40/0.7321 \\
\makecell[l]{DAIN}
& EDVR           & 24.0+20.7                 & 28.01/0.8239  & 26.37/0.7964    & 27.06/0.7895    & 26.01/0.7703 \\
& BasicVSR   & 24.0+6.3                  & 28.00/0.8227  & 26.46/0.7966    & 27.07/0.7890    & 26.23/0.7725 \\
\hline
TimeLens          & EG-VSR                   & 72.2+2.45                  &  28.85/0.8678	      &   27.54/0.8293           &    	28.11/0.8441	           &     27.42/0.8269                        \\
\rrev{CBMNet}          & \rrev{EG-VSR}                   & \rrev{22.2+2.45}                  &  \rrev{29.22/0.8686}	      &   \rrev{28.51/0.8493}           &    	\rrev{28.28/0.8553}	           &     \rrev{27.89/0.8334}                        \\
\hline
\multicolumn{2}{l}{Zooming Slow Mo}             & 11.10                     & 30.69/0.8847  & -/-             & 30.26/0.8821    & -/-          \\
\multicolumn{2}{l}{TMNet}                       & 12.26                     & 30.14/0.8692  & 28.83/0.8514    & 29.41/0.8524    & 28.30/0.8354 \\
\hline
\multicolumn{2}{l}{Video INR-\textit{fixed}}    & 11.31                     & 30.73/0.8850  & -/-             & 30.21/0.8805    & -/-          \\
\multicolumn{2}{l}{Video INR}                   & 11.31                     & 30.26/0.8792  & 29.41/0.8669    & 29.92/0.8746    & 29.27/0.8651 \\
\multicolumn{2}{l}{MoTIF}                          & 12.55                     & 31.04/0.8877  & 30.04/0.8773    & 30.63/0.8839    & 29.82/0.8750 \\
\hline
\multicolumn{2}{l}{HR-INR (Ours)}                                       & 8.27                          & \textbf{ 31.97/0.9298}               & \textbf{ 32.13/0.9371}             &    \textbf{31.26/0.9246}               &  \textbf{31.11/0.9216}             \\
\bottomrule
\end{tabular}
}
}
\end{table*}

\begin{table*}[t!]
\centering
\caption{More quantitative comparisons using PSNR/SSIM on the GoPro dataset. \textbf{Bold} indicates the best performance.\label{tab:comp-gopro-c-stvsr} }
\resizebox{1\linewidth}{!}{
\setlength{\tabcolsep}{0.019\linewidth}{
\begin{tabular}{r|r|c|c|c|c|c|c}
\toprule
\makecell[c]{Temporal\\Scale}
& \makecell[c]{Spatial\\Scale}
& \makecell[c]{Su-SloMo\\ + LIIF }
& \makecell[c]{DAIN \\+ LIIF }
& TMNet
& Video INR
& MoTIF
& Ours \\
\midrule
12$\times$~~~~    & 4$\times$~~~~   & 25.07/0.7491    & 25.14/0.7497 & 26.38/0.7931 & 27.32/0.8141 & 27.77/0.8230 & \textbf{28.87/0.8854} \\
12$\times$~~~~    & 6$\times$~~~~   & 22.91/0.6783    & 22.92/0.6785 & -              & 24.68/0.7358 & 26.78/0.7908 & \textbf{27.14}/\textbf{0.8173} \\
\hline
16$\times$~~~~    & 4$\times$~~~~   & 24.42/0.7296    & 24.20/0.7244 & 24.72/0.7526 & 25.81/0.7739 & 25.98/0.7758 & \textbf{27.29/0.8556} \\
16$\times$~~~~    & 6$\times$~~~~   & 23.28/0.6883    & 22.80/0.6722 & -              & 23.86/0.7123 & 25.34/0.7527 & \textbf{26.09}/\textbf{0.7954} \\
\hline
6$\times$~~~~     & 1$\times$~~~~   & -                 & -              & -              & 32.34/0.9545 & 34.77/0.9696 & \textbf{38.53/0.9735} \\
\rrev{1$\times$}~~~~     & \rrev{4$\times$}~~~~   & -                 & -              & \rrev{33.02/0.9206} & \rrev{32.26/0.9198} & \rrev{\textbf{33.84}/0.9328} &  \rrev{33.51/\textbf{0.9417}}    \\
\bottomrule
\end{tabular}
}
}
\end{table*}

\noindent\textbf{Datasets:} To facilitate a comprehensive comparison between the frame-base \citeps{chen2022videoinr,chen2023motif} and event-guided methods \citeps{lu2023learning,tulyakov2021time,tulyakov2022time,jing2021turning}, we employed two simulated datasets \citeps{Su_Delbracio_Wang_Sapiro_Heidrich_Wang_2017,Nah_Kim_Lee_2017} and two real-world datasets \citeps{tulyakov2022time,Scheerlinck_Rebecq_Stoffregen_Barnes_Mahony_Scaramuzza_2019} for our experiments.
\noindent\textbf{(1)} Adobe240 Dataset~\citeps{Su_Delbracio_Wang_Sapiro_Heidrich_Wang_2017}: This dataset comprises 133 videos, each with a resolution of $720\times 1280$ and a frame rate of 240.
We follow~\citeps{chen2022videoinr,chen2023motif,xu2021temporal} to split this dataset into 100 training, 16 validation, and 17 test sets.
We employed the widely-used event simulation method vid2e~\citeps{vid2e_simlate}, which accounts for real noise distribution, ensuring robust neural network training with enhanced generalization.
In generating the inputs and ground truth of the network, we adopted and extended the previous works~\citeps{chen2023motif,chen2022videoinr} to accommodate multi-frame input.
Specifically, input and ground truth (GT) frames are selected via sliding windows.
We define the window size as $W$, the number of input frames as $N_{in}$, and the interval as $S$.
They interrelate as:
$W = (N_{in} - 1) * (S + 1) + 1$
For instance, with 4 frames input at 7-frame intervals, 25 frames are chosen.
The 1st, 9th, 17th, and 25th frames become the input after down-sampling, while frames 1-25 serve as the GT.
We adopted two strategies in line with prior works~\citeps{chen2022videoinr,chen2023motif}: \textbf{I)} A fixed magnification set at $4\times$ the input resolution, and \textbf{II)} A variable enlargement strategy, wherein the scaling factor is governed by a $\mathcal{U}(1,8)$ distribution.
\textbf{(2)} GoPro dataset \citeps{Nah_Kim_Lee_2017} featuring the same resolution and frame-rate with~ Adobe240 dataset, comprises 22 training and 11 test videos.
Given its compact size, previous studies \citeps{chen2023motif,chen2022videoinr} have primarily employed the test set for quantitative analysis.
We follow this to be consistent with established practice.
\textbf{(3)} BS-ERGB \citeps{tulyakov2022time} recorded using a spectroscope, includes real-world paired events and frames with a $970\times 625$ resolution at 28 fps.
Of the 123 videos in the dataset, 47 are allocated for training, 19 for validation, and 26 for testing.
Each video contains between 100 to 600 frames.
We opted to pre-train our model on the Adobe240 dataset before fine-tuning.
During fine-tuning, we also used perceptual loss \citeps{johnson2016perceptual} with weight $0.1$ to be consistent with previous methods \citeps{tulyakov2021time,tulyakov2022time} for fair comparison.
\textbf{(4)} CED \citeps{Scheerlinck_Rebecq_Stoffregen_Barnes_Mahony_Scaramuzza_2019} is a real-world dataset in VSR. To fairly compare the previous research \citeps{jing2021turning,lu2023learning}, only this data set is used for training for VSR comparison.
It is worth noting that the dataset ALPIX-VSR~\citep{lu2023learning} can also be used for event-guided VSR; however, its events are not stream-based ~\citep{alpsentek}, making it unsuitable for events TPR. Therefore, we use CED as the primary dataset for comparison.

\begin{figure}[t!]
\centering
\includegraphics[width=\linewidth]{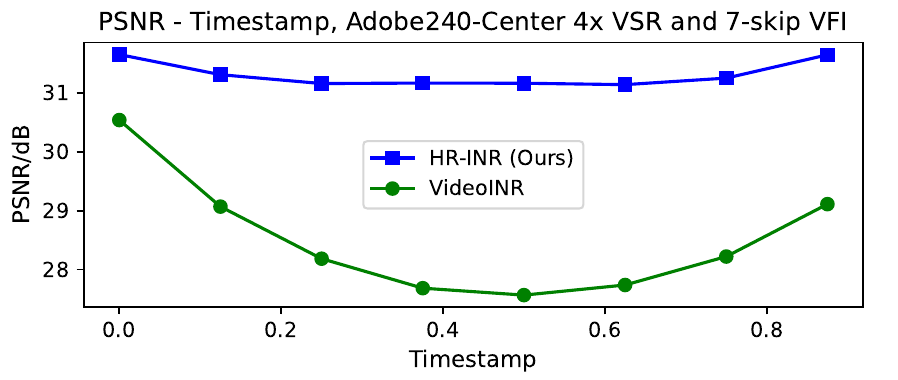}
\caption{Timestamps and corresponding PSNR for each frame during $4\times$ VSR and 7-\textit{skip} VFI on the Adobe240 dataset~\citeps{Su_Delbracio_Wang_Sapiro_Heidrich_Wang_2017}.}
\label{fig:11-PSNR-Time}
\centering
\end{figure}

\noindent \textbf{Implementation Details:}
Our model is trained using Pytorch \citeps{paszke2019pytorch}, employing the Adam optimizer \citeps{kingma2014adam}.
Referring to the VideoINR \citeps{chen2022videoinr}, our training consists of two stages.
\textbf{(1)} Train frame interpolation under a fixed spatial up-sampling ($4\times$),  over 70 epochs, starting with a learning rate of $5e-4$.
\textbf{(2)} Train frame interpolation under random space upsampling rate in $\mathcal{U}(1,8)$, spanning 30 epochs with the learning rate $5e-5$.
We randomly choose 20 frames from a pool of 25 frames as ground truth.
Data augmentation is implemented via \textit{Random Crop}, extracting $512\times 512$ areas from frames and events, with the input resolution dynamically determined by a randomly chosen upsampling ratio $s$.
To optimize memory usage and accelerate speed, we implemented the mixed precision strategy \citeps{micikevicius2017mixed,das2018mixed}.
All experiments are performed on an NVIDIA A800.

\begin{figure*}[t!]
\centering
\includegraphics[width=\linewidth]{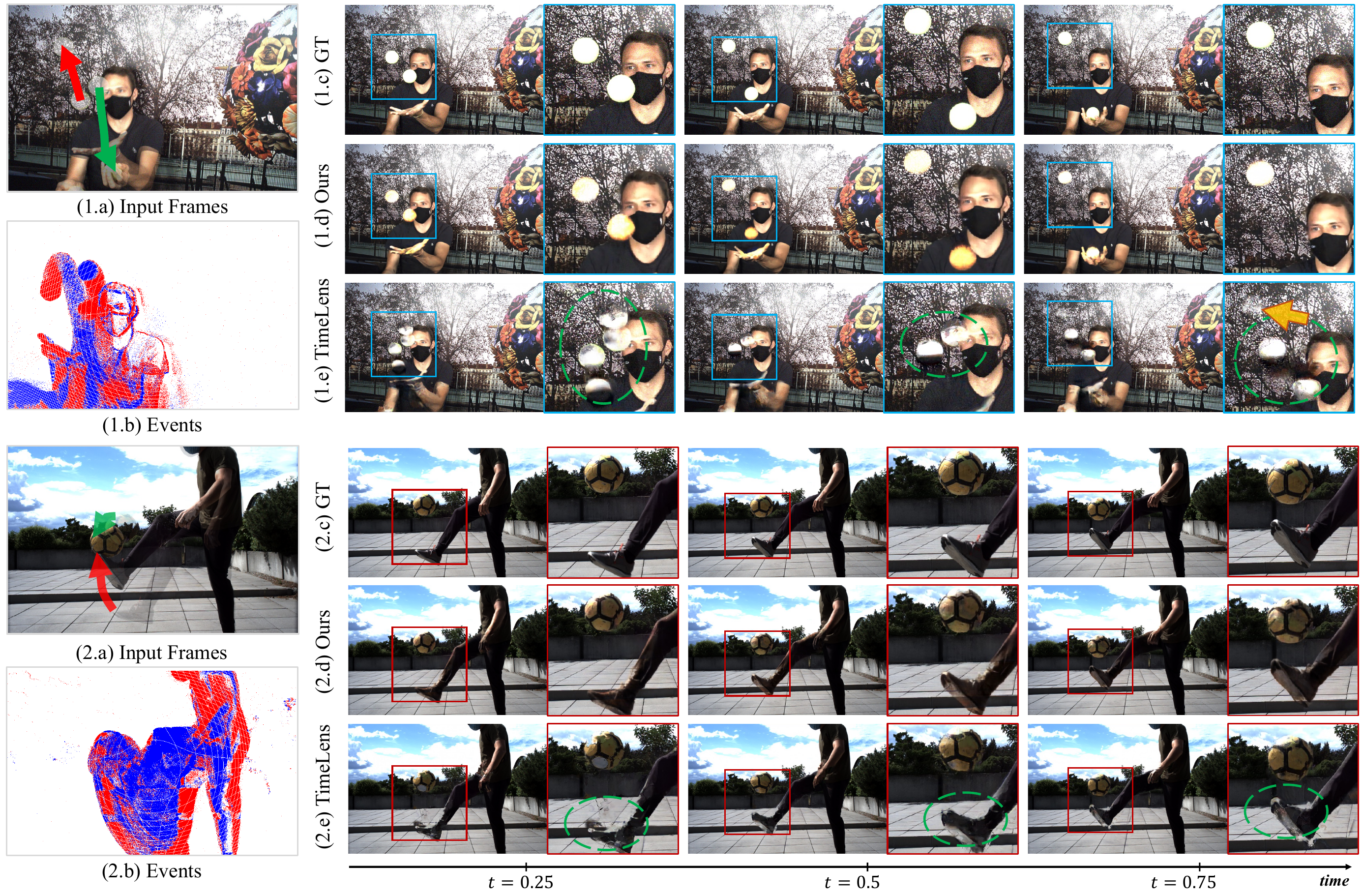}
\caption{
3-\textit{skip} frame interpolation visualization results in BS-ERGB dataset~\citeps{tulyakov2022time}.
Our method accurately captures \ubtxt{local non-linear motion} (\eg, balls in (1.d) and (2.d)), surpassing TimeLens~\citeps{tulyakov2022time}, which exhibits ghosting and holes ({\color{green}green circles}). A {\color{yellow}yellow arrow} shows TimeLens's inaccurate ball positioning.}
\label{fig:15-VFI-in-TimeLensPPDataset-Release}
\centering
\end{figure*}

\noindent\textbf{Evaluation:}
To ensure the fairness, we adopted PSNR \citeps{zhang2018unreasonable}, SSIM \citeps{wang2004image}, and LPIPS \citeps{zhang2018unreasonable} as quantitative evaluation metrics.
For a fair comparison on GoPro and Adobe datasets, we follow previous work \citeps{chen2022videoinr,chen2023motif} by using only the $Y$-channel, while for the BS-ERGB and CED datasets, we employ all three RGB channels.

\subsection{Comparison Experiments\label{sec:comparisino}}

Since our method is the first event-based C-STVSR approach, we compare it with both frame-based methods and cascade methods for C-STVSR, as discussed in Sec. \ref{sec:exp:stsr}.
Additionally, our method also demonstrates superior performance in individual tasks, VFI and VSR. The comparisons for these individual tasks are discussed in Sec. \ref{sec:exp:vfi-vsr}.

\begin{figure*}[t!]
\centering
\includegraphics[width=\linewidth]{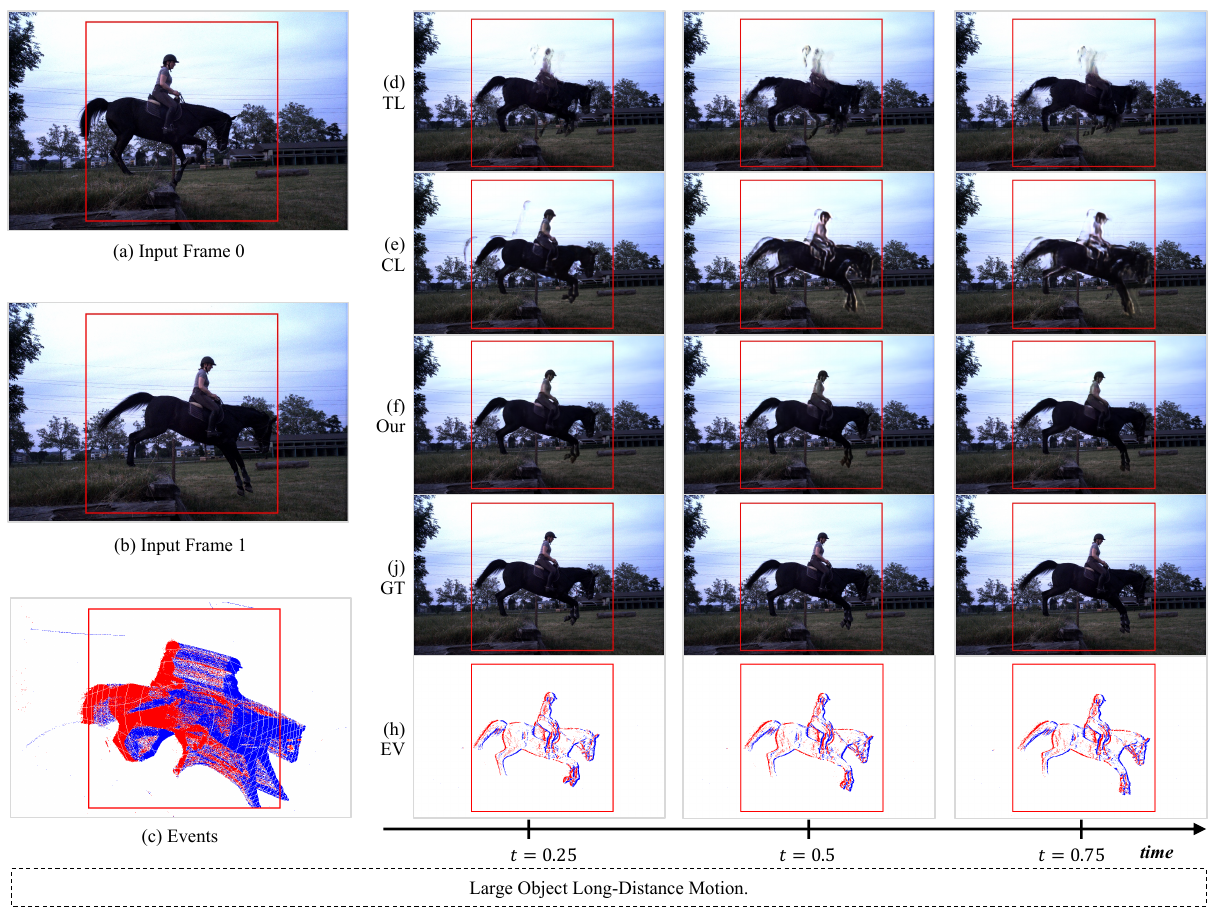}
\caption{Visualization results on a real-world dataset in a \textbf{large object long-distance motion} scene. (a) and (b) represent the input frames, (c) shows the corresponding events, (d) is the result of the TimeLens method (TL), (e) shows the interpolation by CBMNet (CL), (f) presents the result of our method, and (j) is the ground truth (GT), and (h) displays the local events at each time step.\label{fig:20-with-cmbnet-Large}}
\end{figure*}

\subsubsection{Space-time Super-resolution\label{sec:exp:stsr}}
We conduct space-time super-resolution comparison experiments on the Adobe240 and GoPro datasets.
We categorized the comparison methods into three groups.
\textbf{(I)} Cascade methods: VFI methods, \eg, Super SloMo \citeps{jiang2018super}, DAIN \citeps{bao2019depth} and TimeLens~\citep{tulyakov2021time} followed by VSR methods, \eg, EDVR \citeps{wang2019edvr}, BasicVSR \citeps{chan2021basicvsr} and EG-VSR \citep{lu2023learning}.
\textbf{(II)} Fixed STVSR methods: \eg, Zooming Slow-Mo \citeps{xiang2020zooming} and TMNet \citeps{xu2021temporal}.
\textbf{(III)} Frame-based C-STVSR methods: VideoINR \citeps{chen2022videoinr} and MoTIF \citeps{chen2023motif}.

The numerical comparison is presented in Tab.~\ref{tab:comp-gopro-adobe-f-stvsr}.
It is evident that C-STVSR methods consistently outperform cascade and fixed STVSR methods.
Our method achieves the highest performance in both datasets with the smallest model size.
In the GoPro dataset, our method improves the center frame by 0.93 $dB$ and 0.0415 SSIM, and on average by 2.09 $dB$ and 0.0598 SSIM compared to the best method, MoTIF \citeps{chen2023motif}.
Similarly, in the Adobe240 dataset, our method outperforms MoTIF \citeps{chen2023motif} by 0.63 $dB$ and 0.0407 SSIM for the \textit{-center}, and on \textit{-average} by 1.29 $dB$ and 0.0466 SSIM.
It is worth noting that the difference between our method and other methods is more pronounced for the \textit{-average} frames than the \textit{-center} frame.
This observation suggests that our method demonstrates enhanced \ubtxt{temporal~stability}, as shown in Fig.~\ref{fig:11-PSNR-Time}, and superior adaptability across varying interpolation intervals, an advantage not shared by previous methods \citeps{chen2022videoinr,chen2023motif}.

Tab.~\ref{tab:comp-gopro-c-stvsr} presents additional comparative experiments for arbitrary spatial and temporal super-resolution. Our method consistently outperforms other methods, even in extreme space-time upsample scales, such as $16\times$ temporal upscale and $6\times$ spatial upscale.
The visualization results, Fig.~\ref{fig:cover-figure} (c) and Fig.~\ref{fig:7-ADB-VFI-Vis-Release}, also demonstrate that our method effectively models regional nonlinear motion, \eg, the \textbf{wheel rotation} — a capability not achieved by previous method VideoINR \citeps{chen2022videoinr}.
\textit{For more video visualization results, please refer to the Supplementary Materials.}

\noindent\textbf{Stability Across Temporal Sequences:}
Our method demonstrates temporal stability, accurately estimating motion states at each time point during frame interpolation, as shown in Fig. \ref{fig:11-PSNR-Time}.
This is the key reason why our method outperforms previous approaches.
Specifically, whether in the GoPro-\textit{Average} and Adobe-\textit{Average} of Tab.~\ref{tab:comp-gopro-adobe-f-stvsr}, or the 12-\textit{skip} and 16-\textit{skip} tests of Tab.~\ref{tab:comp-gopro-c-stvsr}, our method significantly outperforms previous methods by at least $1.2dB$ with $4\times$ VSR.
\begin{figure*}[t!]
\centering
\includegraphics[width=\linewidth]{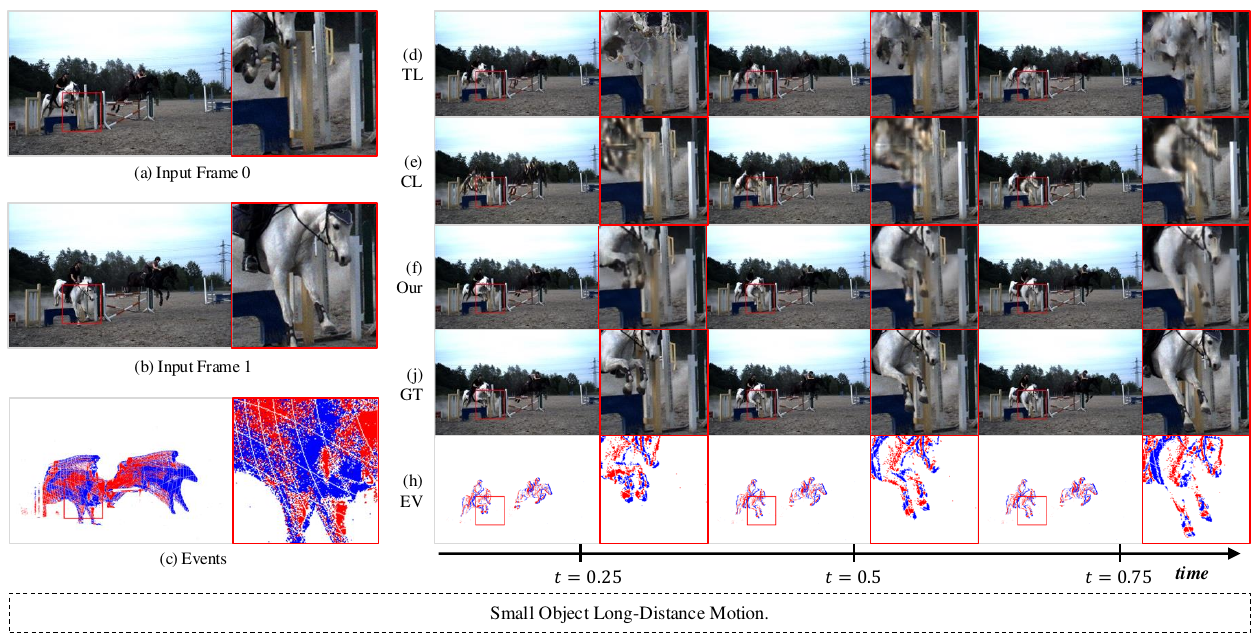}
\caption{Visualization results on a real-world dataset in a \textbf{small object long-distance motion} scene. (a) and (b) represent the input frames, (c) shows the corresponding events, (d) is the result of the TimeLens method (TL), (e) shows the interpolation by CBMNet-Large (CL), (f) presents the result of our method, and (j) is the ground truth (GT), and (h) displays the local events at each time step.}
\label{fig:20-with-cmbnet-Small}
\end{figure*}
\begin{table*}[t!]
\centering
\caption{Temporal super-resolution results, \ie, VFI, on BS-ERGB dataset~\citeps{tulyakov2022time}.}
\resizebox{1\linewidth}{!}{
\setlength{\tabcolsep}{0.015\linewidth}{
\begin{tabular}{l|r|c|ccc|ccc}
\toprule
                                            &               &           & \multicolumn{3}{c}{1-\textit{skip}}                           & \multicolumn{3}{c}{3-\textit{skip}} \\
Methods                                     & Params~($M$)  & Event     & PSNR~$\uparrow$   & SSIM~$\uparrow$   & LPIPS~$\downarrow$    & PSNR~$\uparrow$   & SSIM~$\uparrow$   & LPIPS~$\downarrow$ \\
\hline
\hline
FLAVR~\citeps{kalluri2023flavr}               &  -            & \ding{55} & 25.95             & -                 & 0.086                 & 20.90             & -                 & 0.151 \\
DAIN~\citeps{bao2019depth}                    & 24.0          & \ding{55} & 25.20             & -                 & 0.067                 & 21.40             & -                 & 0.113 \\
Super SloMo~\citeps{jiang2018super}           & 19.8          & \ding{55} & -                 & -                 & -                     & 22.48             & -                 & 0.115 \\
QVI~\citeps{xu2019quadratic}                  & 29.2          & \ding{55} & -                 & -                 & -                     & 23.20             & -                 & 0.110 \\
TimeLens~\citeps{tulyakov2021time}            & 72.2          & \ding{51} & 28.36             & -                 & 0.026                 & 27.58             & -                 & 0.031 \\
TimeLens++~\citeps{tulyakov2022time}          & 53.9          & \ding{51} & 28.56             & -                 & 0.022                 & 27.63             & -                 & 0.026 \\
CBMNet~\citeps{kim2023event}                  & 15.4          & \ding{51} & 29.32             & 0.815             & -                     & 28.46             & 0.806             & -  \\
CBMNet-\textit{Large}~\citeps{kim2023event}   & 22.2          & \ding{51} & 29.43             & 0.816             & -                     & 28.59             & 0.808             & - \\
HR-INR (Our)                            & 8.3           & \ding{51} & \textbf{29.66}    & \textbf{0.828}    & \textbf{0.011}        & \textbf{28.59}    & \textbf{0.814}    & \textbf{0.021}     \\
\bottomrule
\end{tabular}
}
}
\label{tab:vfi-only-bs-ergb-Timelens++}
\end{table*}
Fig.~\ref{fig:11-PSNR-Time} shows the relationship between timestamps and PSNR, validating the greater stability of our method compared to VideoINR~ \citeps{chen2022videoinr}, especially around the $0.5 s$ mark, where VideoINR experiences a notable decrease.
This observation is also reflected in the visualization results of Fig.~\ref{fig:cover-figure},~\ref{fig:7-ADB-VFI-Vis-Release} and the following Fig.~\ref{fig:15-VFI-in-TimeLensPPDataset-Release},~\ref{fig:20-with-cmbnet-Large},~\ref{fig:20-with-cmbnet-Small}.
Whether using real-world or simulated data, our method demonstrates superior temporal consistency, exhibiting fewer artifacts compared to previous methods such as VideoINR~\citeps{chen2022videoinr}, TimeLens~\citeps{tulyakov2021time} and CMB-Net~\citep{kim2023event}.
This also highlights the real-world effectiveness of our method’s temporal stability.
The main reason for this is that previous methods~\citeps{chen2022videoinr, chen2023motif, tulyakov2021time} suffer from instability in motion estimation at timestamps far from the reference frames (at $0$ and $1$ timestamp), which negatively impacts shape and edge estimation for frame interpolation.
In contrast, our method benefits from the intermediate motion details provided by events TPR and the model's ability to capture long temporal sequences, enabling it to maintain higher stability across the temporal sequence and deliver superior performance.

\begin{figure*}[t!]
\centering
\includegraphics[width=\linewidth]{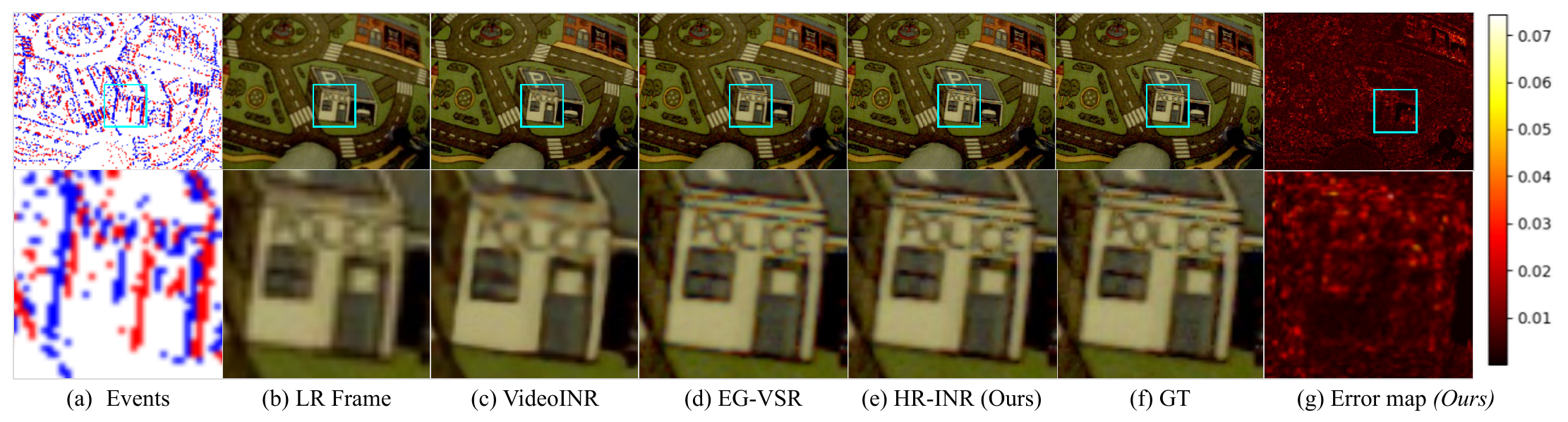}
\caption{
$4\times$ video super-resolution visualization results in CED dataset \citeps{scheerlinck2019ced}.}
\label{fig:6-CED-SR-Vis-Release}
\centering
\end{figure*}
\begin{table*}[t!]
\caption{Spatial super-resolution results on CED~\citeps{scheerlinck2019ced}. * denotes values from pre-trained models.}
\resizebox{1\linewidth}{!}{
\setlength{\tabcolsep}{0.028\linewidth}{
\centering \small
\begin{tabular}{l|r|c|c c| c c}
\toprule
                                        &               &        & \multicolumn{2}{c|}{$4 \times$}  &  \multicolumn{2}{c}{$2 \times$} \\
Methods                                 & Params~($M$)  & Events & PSNR~$\uparrow$  & SSIM~$\uparrow$  &PSNR~$\uparrow$  & SSIM~$\uparrow$\\
\hline
\hline
\makecell[l]{TDAN \citeps{tian2020tdan}}                & 1.97          & \ding{55}     & 27.88 & 0.8231 & 33.74 & 0.9398   \\
\makecell[l]{SOF \citeps{wang2020deep}}                 & 1.00          & \ding{55}     & 27.00 & 0.8050 & 31.84 & 0.9226   \\
\makecell[l]{RBPN \citeps{Haris2019rbpn}}               & 12.18         & \ding{55}     & 29.80 & 0.8975 & 36.66 & 0.9754   \\
\makecell[l]{VideoINR* \citeps{chen2022videoinr}}       & 11.31         & \ding{55}     & 25.53 & 0.7871 & 26.77 & 0.7938  \\
\makecell[l]{E-VSR  \citeps{jing2021turning}}           & 412.42        & \ding{51}     & 30.15 & 0.9052 & 37.32 & 0.9783   \\
\makecell[l]{EG-VSR \citeps{lu2023learning}}            & 2.45          & \ding{51}     & 31.12 & 0.9211 & 38.69 & 0.9771   \\
HR-INR (Our)                            & 8.27          & \ding{51}     & \textbf{32.15} & \textbf{0.9658} & \textbf{42.01} & \textbf{0.9905}        \\
\bottomrule
\end{tabular}
}
}
\label{tab:sr-only-ced-4x}
\end{table*}

\subsubsection{Separate Comparison of Event-based VFI and VSR\label{sec:exp:vfi-vsr}}

We also compared our method with previous approaches in separate VFI \citeps{tulyakov2022time,tulyakov2021time,kim2023event} and VSR \citeps{lu2023learning,jing2021turning} tasks.
The results can be seen in Tab.~\ref{tab:vfi-only-bs-ergb-Timelens++} and Tab.~\ref{tab:sr-only-ced-4x} respectively.

\noindent\textbf{Comparison with Event-based VFI:}
In the VFI task, our method surpasses TimeLens++ \citeps{tulyakov2022time} by 1.1 $dB$ for 1-\textit{skip} and 0.17 $dB$ for 3-\textit{skip} VFI, and CBMNet \citeps{kim2023event} by 0.23 $dB$ for 1-\textit{skip} in Tab.~\ref{tab:vfi-only-bs-ergb-Timelens++}.
Additionally, our model's size is merely $1/7$ and $1/3$ that of TimeLens++ \citeps{tulyakov2022time} and CBMNet \citeps{kim2023event}, respectively.
In the visualization aspect, our approach excels in modeling non-linear motion and long-term dependencies, as evident in Fig.~\ref{fig:15-VFI-in-TimeLensPPDataset-Release}, where the subject is juggling, the motion of small, fast-moving objects (balls) is challenging to capture.
Our method precisely forecasts the positions of small balls (Fig.~\ref{fig:15-VFI-in-TimeLensPPDataset-Release} (1.d)) and the soccer ball (Fig.~\ref{fig:15-VFI-in-TimeLensPPDataset-Release} (2.d)) at intermediate timestamps.
Competing methods exhibit noticeable ghosting and fail to preserve the spatial consistency of the balls.
This demonstrates our model's ability to handle \textbf{high-speed motion} effectively by leveraging its dual-branch feature extraction.

\begin{table*}[t!]
\caption{
Ablation studies in Adobe-\textit{Average}~\citep{Su_Delbracio_Wang_Sapiro_Heidrich_Wang_2017} ($4\times$ and 7-\textit{skip}).
The \dag~ symbol marks the line for comparison with other lines.
\label{tab:ablation-in-ADB-Average-EET}}
\centering
\resizebox{\linewidth}{!}{
\setlength{\tabcolsep}{0.024\linewidth}{
\begin{tabular}{l|cccc|c|ll}
\toprule
Case
& Events
& TPR
& \makecell[c]{Temporal\\ Embedding}
& \makecell[c]{Temporal\\ Dim ($C_t$)}
& \makecell[c]{Input\\ Frames}
& \makecell[c]{PSNR~$\uparrow$}
& \makecell[c]{SSIM~$\uparrow$} \\
\hline
Case\#1             & \pptxt{\ding{55}}        & \pptxt{\ding{55}}     & \textit{Learning}             & 640~~             & 4         & 26.84 (\ssbtxt{-4.27})         & 0.8366 (\ssbtxt{-0.0850})      \\
Case\#2             & \ding{51}                & \pptxt{\ding{55}}     & \textit{Learning}             & 640~~             & 4         & 29.69 (\ssbtxt{-1.41})         & 0.8974 (\ssbtxt{-0.0242})     \\
Case\#3 \dag        & \ding{51}                & \ding{51}             & \textit{Learning}             & 640~~             & 4         & \textbf{31.11}                 & \textbf{0.9216}      \\
\hline
Case\#4             & \ding{51}                & \ding{51}             & \pptxt{\textit{Sinusoid}}     & 640~~             & 4         & 30.42 (\ssbtxt{-0.69})         & 0.9120 (\ssbtxt{-0.0096})      \\
\hline
Case\#5             & \ding{51}                & \ding{51}             & \textit{Learning}             & \pptxt{320}~~     & 4         & 28.44 (\ssbtxt{-2.67})         & 0.8700 (\ssbtxt{-0.0516})       \\
\hline
Case\#6             & \ding{51}                & \ding{51}             & \textit{Learning}             & 640~~             & \pptxt{2} & 30.41 (\ssbtxt{-0.70})         & 0.9151 (\ssbtxt{-0.0065})     \\
Case\#7             & \ding{51}                & \ding{51}             & \textit{Learning}             & 640~~             & \pptxt{3} & 30.72 (\ssbtxt{-0.39})         & 0.9174 (\ssbtxt{-0.0042})-      \\
\bottomrule
\end{tabular}
}
}
\end{table*}

Furthermore, we also explore the performance of our method in scenarios involving motions of objects of different sizes.
Fig.~\ref{fig:20-with-cmbnet-Large}, \ref{fig:20-with-cmbnet-Small} demonstrate the superiority of our method in recovering large-scale and long-distance motions across various real-world scenarios.
These results highlight the effectiveness of our approach in fusing both regional and holistic information to handle complex motion dynamics, while simultaneously reducing ghosting artifacts that are commonly seen in competing methods.
Fig.~\ref{fig:20-with-cmbnet-Large} shows \textit{large object long-distance motion}, such as the horse and rider jumping, our method shows clear advantages. The silhouette of the horse and the rider’s position are reconstructed with remarkable clarity and consistency. In contrast, competing methods introduce significant blurring and fail to preserve the integrity of the subject’s shape, revealing their limitations in handling large-scale motion over time.
Fig.~\ref{fig:20-with-cmbnet-Small} shows \textit{small object long-distance motion}. Our method excels in reconstructing fine-grained motion details, such as the movement of the horse's legs. Unlike CMB-Net and TimeLens, which either fail to capture these intricate details or introduce substantial motion blur, our approach accurately restores the distinct positions of the horse’s hooves.
This ability arises from the integration of local motion features captured by the regional branch, along with the long-term temporal dependencies modeled by the holistic branch.
Further analysis can be found in Sec.~\ref{sec:analytical_experiments}.

Across all scenarios, our method generates outputs with fewer ghosting artifacts. The combination of regional and holistic features allows our approach to capture both short-term and long-term motion dependencies, resulting in more realistic and temporally coherent video outputs. These results validate the robustness and generalizability of our method across diverse motion patterns and scales.

\noindent\textbf{Comparison with Event-based VSR:}
For the VSR task, our method outperforms EG-VSR \citeps{lu2023learning} by 1.03 $dB$ and 0.447 SSIM for 4 $\times$ super-resolution, and by 3.02 $dB$ and 0.0134 SSIM for 2 $\times$ super-resolution. As shown in Fig. 10, our method produces visually sharper results compared to EG-VSR and other baseline methods. In Fig. ~\ref{fig:6-CED-SR-Vis-Release}, we present video super-resolution visualization results on the CED dataset \citeps{scheerlinck2019ced}, where our approach effectively restores fine details, such as the text on the building and edges around structures, which are blurred or distorted in competing methods. The corresponding error map (Fig. ~\ref{fig:6-CED-SR-Vis-Release}~(g)) further demonstrates our method's reduced errors, especially around high-contrast regions.

\begin{figure*}[t!]
\centering
\includegraphics[width=\linewidth]{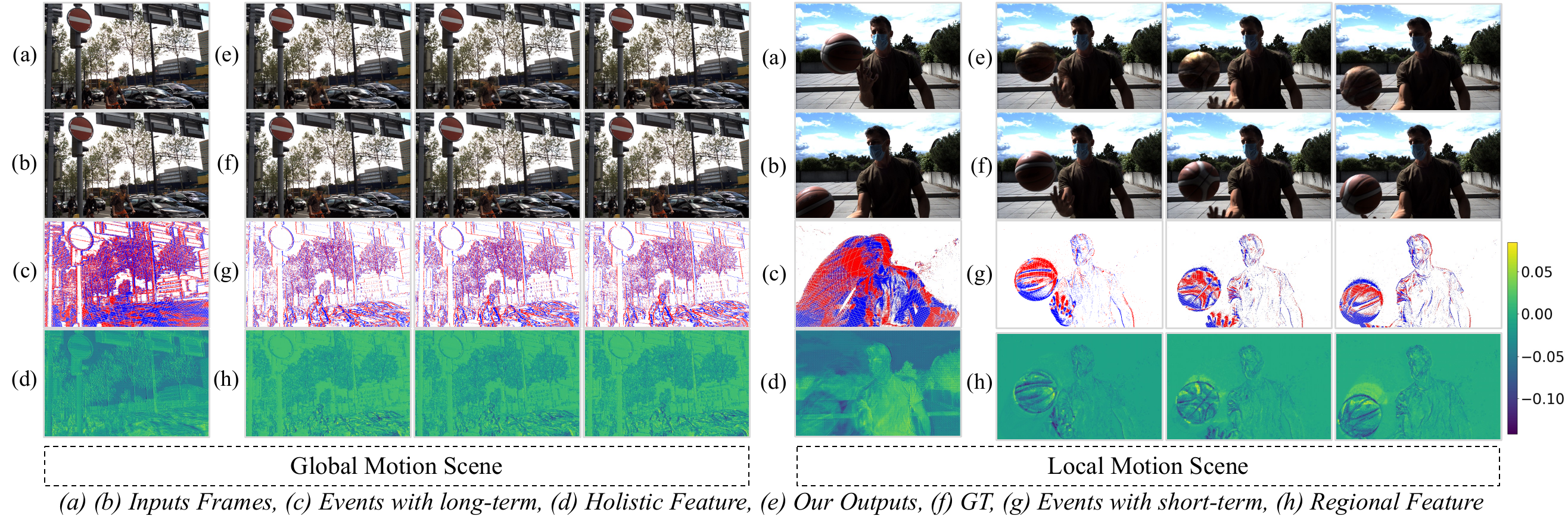}
\caption{
\rrev{Feature visualization on real data~\citeps{tulyakov2022time}: \textbf{(f)} shows the holistic feature, $F^g$, derived from multiple frames and events; \textbf{(g)} depict the regional features ($F_t^g$), highlighting the capability to capture local motion.}}
\label{fig:14-FeatureVisualization}
\centering
\end{figure*}

\subsection{Ablation and Analytical Studies\label{sec:analytical_experiments}}
Ablation and analytical studies conducted on the Adobe240 \citeps{Su_Delbracio_Wang_Sapiro_Heidrich_Wang_2017} and BS-ERGB \citeps{tulyakov2022time} datasets unveiled several critical insights.
On the Adobe240 dataset, we executed simultaneous 7-\textit{skip} VFI and $4\times$ VSR tests, as shown in Tab.~\ref{tab:ablation-in-ADB-Average-EET} and Tab.~\ref{tab:ablation-in-ADB-Average-ETPR}, while on the BS-ERGB dataset, 1-\textit{skip} and 3-\textit{skip} VFI were performed in Tab.~\ref{tab:ablation-vfi-only-bs-ergb-Timelens++}.

\begin{figure*}[t!]
\centering
\includegraphics[width=\linewidth]{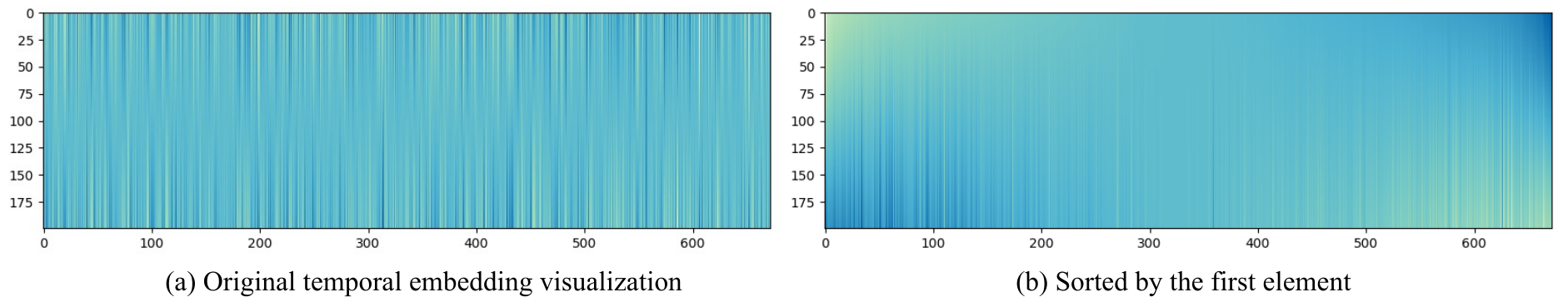}
\caption{Visualization of Temporal Embedding. Figure (a) shows the original visualization of Temporal Embedding, while figure (b) displays the results after sorting. The sorting is based on the size of the first element, arranged in ascending order. We use MLP decoding, where the order is not crucial. However, to more clearly demonstrate the outcomes of Temporal Embedding learning, we have chosen to present the sorted results.}
\label{fig:11-TemporalVisualizationOriginalSorted}
\centering
\end{figure*}
\begin{table*}[t!]
\caption{Ablation studies for TPR levels and moments in Adobe-\textit{Average}~\cite{Su_Delbracio_Wang_Sapiro_Heidrich_Wang_2017} ($4\times$ and 7-\textit{skip}).
The \dag~ symbol marks the line for comparison with other lines.
"Captured Moment" refers to the temporal resolution of the last layer of the TPR, which is calculated by Eq.~\ref{eq:time_granularity}.
\label{tab:ablation-in-ADB-Average-ETPR}}
\centering\small
\resizebox{1\linewidth}{!}{
\setlength{\tabcolsep}{0.019\linewidth}{
\begin{tabular}{l|ccl|ll}
\toprule
Case & TPR Level ($L$) & TPR Moments ($M_p$)
& Captured Moment
& PSNR~$\uparrow$ & SSIM~$\uparrow$ \\
\hline
Case\#1         & 3          & 3    & 1 / 81 & 29.93 (\ssbtxt{-1.18})                & 0.9011 (\ssbtxt{-0.0205})      \\
Case\#2         & 5          & 3    & 1 / 729 & 30.32 (\ssbtxt{-0.79})                & 0.9165 (\ssbtxt{-0.0051})    \\
Case\#3         & 7          & 3    & 1 / 6561 & 30.78 (\ssbtxt{-0.33})                & 0.9187 (\ssbtxt{-0.0029})    \\
Case\#4 \dag    & 7          & 9    & 1 / 19683 & \textit{\textbf{31.11}}               & \textit{\textbf{0.9216}}      \\
Case\#5         & 7          & 18   & 1 / 39366 & \textbf{31.18} (\sgntxt{+0.07})       & \textbf{0.9228} (\sgntxt{+0.0012})     \\
\bottomrule
\end{tabular}
}
}
\end{table*}

\begin{table*}[t!]
\caption{Ablation in BS-ERGB (TimeLens++) dataset~\cite{tulyakov2022time}.\label{tab:ablation-vfi-only-bs-ergb-Timelens++}}
\small\centering
\resizebox{1\linewidth}{!}{
\setlength{\tabcolsep}{0.020\linewidth}{
\begin{tabular}{c|l|l|l|l|l|l}
\toprule
               & \multicolumn{3}{c|}{1-\textit{skip}}& \multicolumn{3}{c}{3-\textit{skip}} \\
TPR  &
\makecell[c]{PSNR~$\uparrow$}  &
\makecell[c]{SSIM~$\uparrow$} &
\makecell[c]{LPIPS~$\downarrow$} &
\makecell[c]{PSNR~$\uparrow$}  &
\makecell[c]{SSIM~$\uparrow$} &
\makecell[c]{LPIPS~$\downarrow$} \\
\hline
\ding{55}
& 28.25                                 & 0.8187                                & 0.018
& 26.65                                 & 0.7867                                & 0.039     \\
\ding{51}
& \textbf{29.66} (\sgntxt{+1.41})       & \textbf{0.8281} (\sgntxt{+0.0094})    & \textbf{0.011} (\sgntxt{-0.007})
& \textbf{28.59} (\sgntxt{+1.94})       & \textbf{0.8140} (\sgntxt{+0.0273})    & \textbf{0.021} (\sgntxt{-0.018})    \\
\bottomrule
\end{tabular}
}
}
\end{table*}

\noindent\textbf{Events Gain}:
Tab.~\ref{tab:ablation-in-ADB-Average-EET}-\textit{Case\#1} shows that with events input replaced as zero and unchanged network architecture, PSNR and SSIM significantly drop, highlighting the importance of events for temporal motion learning.
Adding events alone substantially raised PSNR by $2.85dB$ and SSIM by $0.06$.

\noindent\textbf{Event TPR Analysis:}
Incorporating event TPR significantly enhances performance, with both PSNR and SSIM improving as the layer count increases, peaking at seven layers, as shown in Tab.~\ref{tab:ablation-in-ADB-Average-EET} and Tab.~\ref{tab:ablation-in-ADB-Average-ETPR}.
Specifically, in Tab.~\ref{tab:ablation-in-ADB-Average-ETPR}, as the TRP Level $L$ increases, the moments captured by the TPR become more precise.
For example, when the TRP Level rises from 3 in Tab.~\ref{tab:ablation-in-ADB-Average-ETPR}-\textit{Case\#1}, to 7 in Tab.~\ref{tab:ablation-in-ADB-Average-ETPR}-\textit{Case\#4}, the PSNR shows an increase of approximately 1.18.
Moreover, the model’s performance further improves with the increase in TPR Moments $M_p$. However, it is important to note that when both $L$ and $M_p$ are relatively high, the performance improvement tends to plateau.
Additionally, the TPR enhancement is also evident on the BS-ERGB dataset, as shown in Tab.~\ref{tab:ablation-vfi-only-bs-ergb-Timelens++}, where it yields an increase of $0.92$ dB for 1-\textit{skip} and an improvement of $1.13$ dB for 3-\textit{skip}.

\noindent\textit{\textbf{Analysis of Event TPR in Visualization}:}
Building upon the performance analysis of TPR, we now examine its effectiveness in visualization. As shown in Fig.~\ref{fig:14-FeatureVisualization}, TPR’s regional features emphasize localized motion, demonstrating its ability to capture fine-grained, short-term dynamics. This capability highlights TPR as a novel event representation for millisecond-scale motion at high temporal precision.
For instance, the regional features (Fig.\ref{fig:14-FeatureVisualization}\textbf{(h)}) focus on localized, short-term motion, while the holistic features (Fig.\ref{fig:14-FeatureVisualization}\textbf{(d)}) capture the broader, long-term context. Together, these two types of features allow our method to effectively separate dynamic motion from static background information.
In the \textit{Global Motion Scene}, as shown in Fig.~\ref{fig:14-FeatureVisualization}, where vehicles move against a stationary urban background, the regional features capture the intricate, millisecond-level motion of the vehicles, while the holistic features retain the overall structure of the scene, including static elements such as traffic signs and buildings. This dual representation enables our method to handle complex dynamic scenes seamlessly, where fast-moving objects coexist with stationary elements.
In the \textit{Local Motion Scene}, as shown in Fig.~\ref{fig:14-FeatureVisualization}, the visualization of a basketball player dribbling the ball reveals that the holistic features primarily encode static background details like trees and stationary objects, while the regional features focus on the fast, dynamic motion of the basketball. This demonstrates TPR’s strength in isolating short-term motion, even for high-speed moving objects.
These visualizations provide strong evidence of TPR’s capability to capture fine-grained, short-term motion through its hierarchical structure. By focusing on temporal granularity, event data complements frame-based information and enables the extraction of rich local motion features.
We also encourage readers to refer to the \textbf{\textit{supplementary videos}} for further demonstrations, where these cases are presented in more detail, showcasing how TPR integrates both event and frame data to effectively handle complex motion patterns. This visualization underscores the unique advantages of TPR, combining regional and holistic information to enable robust spatiotemporal super-resolution.

\noindent\textbf{Time Embedding Method:}
Tab.~\ref{tab:ablation-in-ADB-Average-EET}-\textit{Case\#3\#4} presents the results for sinusoid embeddings. The learning-based methods perform better due to their superior ability to capture high-frequency information \citeps{ramasinghe2023learnable,attal2022learning}.
This advantage is further demonstrated in Fig.~\ref{fig:11-TemporalVisualizationOriginalSorted}, which visualizes the time embedding features. The learning-based embeddings not only effectively capture periodic positional representations but also offer a more comprehensive expression of exposure information.

\noindent\textbf{Temporal Dimension Impact:}
The INR temporal dimension significantly influences performance. Lowering the dimensions from 640 to 320 degrades performance in Tab.~\ref{tab:ablation-in-ADB-Average-EET}-\textit{Case\#3\#5}, suggesting a reduction in temporal detail capture. Conversely, expanding the dimension to 960 poses instability risks (\eg, $NAN$ errors).
This highlights the need to balance dimensionality and training stability.

\begin{figure}[t!]
    \centering
    \includegraphics[width=\linewidth]{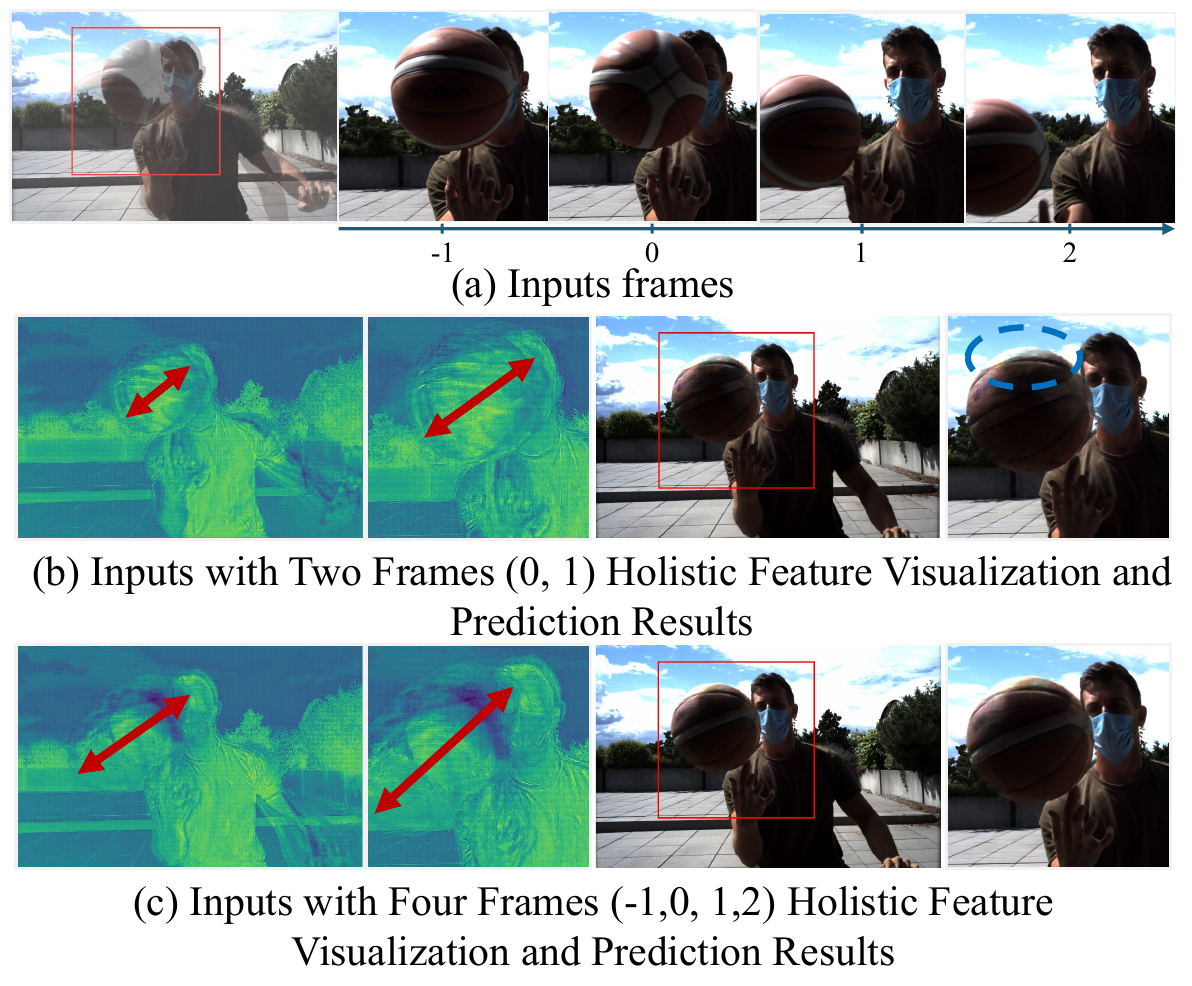}
    \caption{Visualization of Holistic Features for different frames as input.}
    \label{fig:R2-Holistic-Feature-Frames}
\end{figure}

\noindent\textbf{Analysis of Input Frames and Long-distance Modeling:} Tab.~\ref{tab:ablation-in-ADB-Average-EET}-\textit{Case\#3\#6\#7} illustrates the impact of varying input frame counts on the final results. We observed a performance decrease of 0.70 $dB$ and 0.39 $dB$ when inputting two and three frames, respectively, compared to four frames. This indicates a clear advantage of multi-frame inputs in modeling longer-term dependencies. Moreover, even with two frames, our method also outperforms previous works \citeps{chen2022videoinr,chen2023motif}.
Fig. \ref{fig:R2-Holistic-Feature-Frames} compares the outputs of two models: one using two input frames (0, 1) and another using four input frames (-1, 0, 1, 2).
With two input frames, the model struggles to capture complex motion, as seen in the less accurate predictions of the basketball’s trajectory in Fig.{\ref{fig:R2-Holistic-Feature-Frames}}~(b). In contrast, the four-frame model leverages additional temporal information to model longer-term motion, producing smoother and more accurate results, as shown in Fig.\ref{fig:R2-Holistic-Feature-Frames}~(c). The holistic feature visualizations further highlight the richer temporal dependencies captured by the four-frame model, allowing it to handle challenging motion scenarios more effectively.
This analysis underscores the strength of our method in utilizing \textbf{multi-frame inputs} to improve temporal coherence and enhance interpolation quality, particularly for long-distance or rapid motion.

\begin{figure}[t!]
\centering
\includegraphics[width=\linewidth]{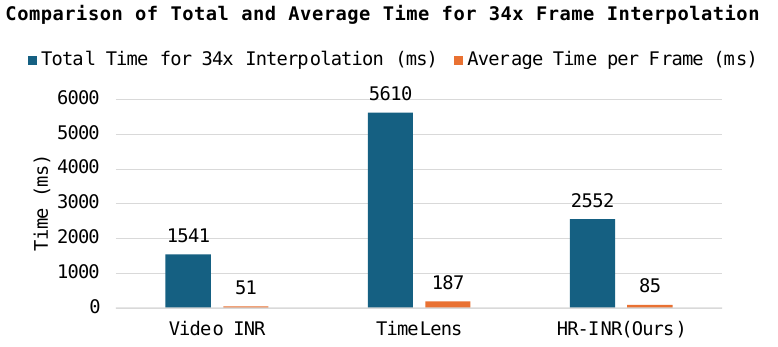}
\caption{Comparison of total and average time for $34 \times$ frame interpolation by different methods. Our method takes less time than TimeLens~\citeps{tulyakov2021time}, but slightly more time than VideoINR~\citeps{chen2022videoinr}.}
\label{fig:19-interpolation-time-difference-method}
\centering
\end{figure}

\noindent\textbf{Inference time analysis}
In Fig.~\ref{fig:11-TemporalVisualizationOriginalSorted}, we analyze the inference times of three different methods. Both our method and VideoINR achieve an average frame time of less than 100 $ms$ for $34 \times$ frame interpolation. In contrast, TimeLens~\citeps{tulyakov2021time} has an average frame time of 187 $ms$, which is more than double that of our method. The tests were conducted on a high-performance computer, and each method was tested 30 times, with the final inference time being the average of these 30 trials.

\noindent\textbf{Bad case analysis}
We have observed that our method has certain limitations in some cases (Fig.~\ref{fig:15-VFI-in-TimeLensPPDataset-Release}).
For example, when restoring color information, although our model can accurately reconstruct the contours of objects, the color information is often distorted or missing. This issue primarily arises due to the lack of color information in the event stream. We believe that with future advancements in color event technology, this problem will be effectively addressed.

\vsection{Conclusion}
Our work introduced the first event-guided continuous space-time video super-resolution method. The main contributions are:

\begin{itemize}
\item \textbf{Event temporal pyramid representation} for capturing short-term dynamic motion;
\item \textbf{A feature extraction process} combining holistic and regional features to manage motion dependencies;
\item \textbf{A spatiotemporal decoding process based on implicit neural representation,} avoiding traditional optical flow and achieving stable frame interpolation through temporal-spatial embedding.
\end{itemize}

Furthermore, the unique combination of event data with temporal pyramid representation allows our method to handle both fast-moving and stationary objects seamlessly, making it highly effective in real-world scenarios with complex motion. Our approach also outperforms previous methods in terms of stability, both in temporal and spatial resolution, as demonstrated by comprehensive experiments on multiple datasets. By integrating local and global motion features, our method offers enhanced adaptability, overcoming challenges such as ghosting artifacts and motion blur, commonly encountered in prior techniques. This not only sets a new benchmark in video super-resolution but also paves the way for more accurate and efficient event-based video processing in dynamic environments. In the future, we plan to extend this method to other fields, such as deblurring, to further enhance its applicability.

\begin{acknowledgements}
This work was supported in part by the National Key R\&D Program of China (Grant No.2023YFF0725001), in part by the National Natural Science Foundation of China (Grant No.92370204), in part by the guangdong Basic and Applied Basic Research Foundation (Grant No.2023B1515120057). in part by Guangzhou-HKUST(GZ) Joint Funding Program (Grant No.2023A03J0008), Education Bureau of Guangzhou Municipality.
\end{acknowledgements}
{\small
\bibliographystyle{spbasic}
\bibliography{egbib}

\begin{thebibliography}{88}
\providecommand{\natexlab}[1]{#1}
\providecommand{\url}[1]{{#1}}
\providecommand{\urlprefix}{URL }
\expandafter\ifx\csname urlstyle\endcsname\relax
  \providecommand{\doi}[1]{DOI~\discretionary{}{}{}#1}\else
  \providecommand{\doi}{DOI~\discretionary{}{}{}\begingroup
  \urlstyle{rm}\Url}\fi
\providecommand{\eprint}[2][]{\url{#2}}

\bibitem[{{Alpsentek}(2024)}]{alpsentek}
{Alpsentek} (2024) Alpix-eiger product:. Online:
  \url{https://alpsentek.com/product},
  \urlprefix\url{https://alpsentek.com/product}, accessed: 2024-12-19

\bibitem[{Attal et~al(2022)Attal, Huang, Zollh{\"o}fer, Kopf, and
  Kim}]{attal2022learning}
Attal B, Huang JB, Zollh{\"o}fer M, Kopf J, Kim C (2022) Learning neural light
  fields with ray-space embedding. In: Proceedings of the IEEE/CVF Conference
  on Computer Vision and Pattern Recognition, pp 19819--19829

\bibitem[{Ba et~al(2016)Ba, Kiros, and Hinton}]{ba2016layer}
Ba JL, Kiros JR, Hinton GE (2016) Layer normalization. arXiv preprint
  arXiv:160706450

\bibitem[{Bao et~al(2019{\natexlab{a}})Bao, Lai, Ma, Zhang, Gao, and
  Yang}]{bao2019depth}
Bao W, Lai WS, Ma C, Zhang X, Gao Z, Yang MH (2019{\natexlab{a}}) Depth-aware
  video frame interpolation. In: Proceedings of the IEEE/CVF Conference on
  Computer Vision and Pattern Recognition, pp 3703--3712

\bibitem[{Bao et~al(2019{\natexlab{b}})Bao, Lai, Zhang, Gao, and
  Yang}]{bao2019memc}
Bao W, Lai WS, Zhang X, Gao Z, Yang MH (2019{\natexlab{b}}) Memc-net: Motion
  estimation and motion compensation driven neural network for video
  interpolation and enhancement. IEEE transactions on pattern analysis and
  machine intelligence 43(3):933--948

\bibitem[{Cao et~al(2023)Cao, Wang, Xian, Li, Ni, Pi, Zhang, Zhang, Timofte,
  and Van~Gool}]{cao2023ciaosr}
Cao J, Wang Q, Xian Y, Li Y, Ni B, Pi Z, Zhang K, Zhang Y, Timofte R, Van~Gool
  L (2023) Ciaosr: Continuous implicit attention-in-attention network for
  arbitrary-scale image super-resolution. In: Proceedings of the IEEE/CVF
  Conference on Computer Vision and Pattern Recognition, pp 1796--1807

\bibitem[{Chan et~al(2021)Chan, Wang, Yu, Dong, and Loy}]{chan2021basicvsr}
Chan KC, Wang X, Yu K, Dong C, Loy CC (2021) Basicvsr: The search for essential
  components in video super-resolution and beyond. In: Proceedings of the
  IEEE/CVF Conference on Computer Vision and Pattern Recognition

\bibitem[{Chan et~al(2022{\natexlab{a}})Chan, Zhou, Xu, and
  Loy}]{chan2022basicvsr++}
Chan KC, Zhou S, Xu X, Loy CC (2022{\natexlab{a}}) Basicvsr++: Improving video
  super-resolution with enhanced propagation and alignment. In: Proceedings of
  the IEEE/CVF Conference on Computer Vision and Pattern Recognition, pp
  5972--5981

\bibitem[{Chan et~al(2022{\natexlab{b}})Chan, Zhou, Xu, and
  Loy}]{chan2021basicvsr++}
Chan KC, Zhou S, Xu X, Loy CC (2022{\natexlab{b}}) Basicvsr++: Improving video
  super-resolution with enhanced propagation and alignment. In: Proceedings of
  the IEEE/CVF Conference on Computer Vision and Pattern Recognition

\bibitem[{Chen et~al(2023{\natexlab{a}})Chen, Xu, Hong, Tsai, Kuo, and
  Lee}]{chen2023cascaded}
Chen HW, Xu YS, Hong MF, Tsai YM, Kuo HK, Lee CY (2023{\natexlab{a}}) Cascaded
  local implicit transformer for arbitrary-scale super-resolution. In:
  Proceedings of the IEEE/CVF Conference on Computer Vision and Pattern
  Recognition, pp 18257--18267

\bibitem[{Chen et~al(2025)Chen, Feng, Cai, Wang, Burner, Yuan, Fermuller,
  Metzler, and Aloimonos}]{chen2025repurposing}
Chen J, Feng BY, Cai H, Wang T, Burner L, Yuan D, Fermuller C, Metzler CA,
  Aloimonos Y (2025) Repurposing pre-trained video diffusion models for
  event-based video interpolation. In: Proceedings of the Computer Vision and
  Pattern Recognition Conference, pp 12456--12466

\bibitem[{Chen et~al(2021)Chen, Liu, and Wang}]{chen2021learning}
Chen Y, Liu S, Wang X (2021) Learning continuous image representation with
  local implicit image function. In: Proceedings of the IEEE/CVF Conference on
  Computer Vision and Pattern Recognition, pp 8628--8638

\bibitem[{Chen et~al(2023{\natexlab{b}})Chen, Chen, Lin, and
  Peng}]{chen2023motif}
Chen YH, Chen SC, Lin YY, Peng WH (2023{\natexlab{b}}) Motif: Learning motion
  trajectories with local implicit neural functions for continuous space-time
  video super-resolution. In: Proceedings of the IEEE/CVF International
  Conference on Computer Vision, pp 23131--23141

\bibitem[{Chen et~al(2022)Chen, Chen, Liu, Xu, Goel, Wang, Shi, and
  Wang}]{chen2022videoinr}
Chen Z, Chen Y, Liu J, Xu X, Goel V, Wang Z, Shi H, Wang X (2022) Videoinr:
  Learning video implicit neural representation for continuous space-time
  super-resolution. In: Proceedings of the IEEE/CVF Conference on Computer
  Vision and Pattern Recognition, pp 2047--2057

\bibitem[{Cheng and Chen(2020)}]{cheng2020video}
Cheng X, Chen Z (2020) Video frame interpolation via deformable separable
  convolution. In: Proceedings of the AAAI Conference on Artificial
  Intelligence, vol~34, pp 10607--10614

\bibitem[{Das et~al(2018)Das, Mellempudi, Mudigere, Kalamkar, Avancha,
  Banerjee, Sridharan, Vaidyanathan, Kaul, Georganas et~al}]{das2018mixed}
Das D, Mellempudi N, Mudigere D, Kalamkar D, Avancha S, Banerjee K, Sridharan
  S, Vaidyanathan K, Kaul B, Georganas E, et~al (2018) Mixed precision training
  of convolutional neural networks using integer operations. arXiv preprint
  arXiv:180200930

\bibitem[{Delbracio et~al(2021)Delbracio, Kelly, Brown, and
  Milanfar}]{delbracio2021mobile}
Delbracio M, Kelly D, Brown MS, Milanfar P (2021) Mobile computational
  photography: A tour. Annual Review of Vision Science 7:571--604

\bibitem[{Dutta et~al(2021)Dutta, Shah, and Mittal}]{dutta2021efficient}
Dutta S, Shah NA, Mittal A (2021) Efficient space-time video super resolution
  using low-resolution flow and mask upsampling. In: Proceedings of the
  IEEE/CVF Conference on Computer Vision and Pattern Recognition, pp 314--323

\bibitem[{Gallego et~al(2020)Gallego, Delbr{\"u}ck, Orchard, Bartolozzi, Taba,
  Censi, Leutenegger, Davison, Conradt, Daniilidis et~al}]{gallego2020event}
Gallego G, Delbr{\"u}ck T, Orchard G, Bartolozzi C, Taba B, Censi A,
  Leutenegger S, Davison AJ, Conradt J, Daniilidis K, et~al (2020) Event-based
  vision: A survey. IEEE Transactions on Pattern Analysis and Machine
  Intelligence 44(1):154--180

\bibitem[{Gehrig et~al(2020)Gehrig, Gehrig, Hidalgo-Carrio, and
  Scaramuzza}]{vid2e_simlate}
Gehrig D, Gehrig M, Hidalgo-Carrio J, Scaramuzza D (2020) Video to events:
  Recycling video datasets for event cameras. In: 2020 IEEE/CVF Conference on
  Computer Vision and Pattern Recognition (CVPR),
  \doi{10.1109/cvpr42600.2020.00364},
  \urlprefix\url{http://dx.doi.org/10.1109/cvpr42600.2020.00364}

\bibitem[{Geng et~al(2022)Geng, Liang, Ding, and Zharkov}]{geng2022rstt}
Geng Z, Liang L, Ding T, Zharkov I (2022) Rstt: Real-time spatial temporal
  transformer for space-time video super-resolution. In: Proceedings of the
  IEEE/CVF Conference on Computer Vision and Pattern Recognition, pp
  17441--17451

\bibitem[{Han et~al(2021)Han, Yang, Zhou, Xu, and Shi}]{Han2021EvIntSR}
Han J, Yang Y, Zhou C, Xu C, Shi B (2021) Evintsr-net: Event guided multiple
  latent frames reconstruction and super-resolution. In: Proceedings of the
  IEEE/CVF International Conference on Computer Vision, pp 4882--4891

\bibitem[{Haris et~al(2019)Haris, Shakhnarovich, and Ukita}]{Haris2019rbpn}
Haris M, Shakhnarovich G, Ukita N (2019) Recurrent back-projection network for
  video super-resolution. In: 2019 IEEE/CVF Conference on Computer Vision and
  Pattern Recognition (CVPR), pp 3892--3901

\bibitem[{Haris et~al(2020)Haris, Shakhnarovich, and Ukita}]{haris2020space}
Haris M, Shakhnarovich G, Ukita N (2020) Space-time-aware multi-resolution
  video enhancement. In: Proceedings of the IEEE/CVF Conference on Computer
  Vision and Pattern Recognition, pp 2859--2868

\bibitem[{He et~al(2022)He, You, Qiao, Jia, Zhang, Wang, Lu, Wang, and
  Liao}]{he2022timereplayer}
He W, You K, Qiao Z, Jia X, Zhang Z, Wang W, Lu H, Wang Y, Liao J (2022)
  Timereplayer: Unlocking the potential of event cameras for video
  interpolation. In: Proceedings of the IEEE/CVF Conference on Computer Vision
  and Pattern Recognition, pp 17804--17813

\bibitem[{Isobe et~al(2020)Isobe, Li, Jia, Yuan, Slabaugh, Xu, Li, Wang, and
  Tian}]{isobe2020video}
Isobe T, Li S, Jia X, Yuan S, Slabaugh G, Xu C, Li YL, Wang S, Tian Q (2020)
  Video super-resolution with temporal group attention. In: Proceedings of the
  IEEE/CVF Conference on Computer Vision and Pattern Recognition, pp 8008--8017

\bibitem[{Jiang et~al(2018)Jiang, Sun, Jampani, Yang, Learned-Miller, and
  Kautz}]{jiang2018super}
Jiang H, Sun D, Jampani V, Yang MH, Learned-Miller E, Kautz J (2018) Super
  slomo: High quality estimation of multiple intermediate frames for video
  interpolation. In: Proceedings of the IEEE Conference on Computer Vision and
  Pattern Recognition, pp 9000--9008

\bibitem[{Jing et~al(2021)Jing, Yang, Wang, Song, and Tao}]{jing2021turning}
Jing Y, Yang Y, Wang X, Song M, Tao D (2021) Turning frequency to resolution:
  Video super-resolution via event cameras. In: Proceedings of the IEEE/CVF
  Conference on Computer Vision and Pattern Recognition, pp 7772--7781

\bibitem[{Johnson et~al(2016)Johnson, Alahi, and
  Fei-Fei}]{johnson2016perceptual}
Johnson J, Alahi A, Fei-Fei L (2016) Perceptual losses for real-time style
  transfer and super-resolution. In: Computer Vision--ECCV 2016: 14th European
  Conference, Amsterdam, The Netherlands, October 11-14, 2016, Proceedings,
  Part II 14, Springer, pp 694--711

\bibitem[{Kalluri et~al(2023)Kalluri, Pathak, Chandraker, and
  Tran}]{kalluri2023flavr}
Kalluri T, Pathak D, Chandraker M, Tran D (2023) Flavr: Flow-agnostic video
  representations for fast frame interpolation. In: Proceedings of the IEEE/CVF
  Winter Conference on Applications of Computer Vision, pp 2071--2082

\bibitem[{Kim et~al(2020)Kim, Oh, and Kim}]{Kim_Oh_Kim_2020}
Kim SY, Oh J, Kim M (2020) Fisr: Deep joint frame interpolation and
  super-resolution with a multi-scale temporal loss. Proceedings of the AAAI
  Conference on Artificial Intelligence p 11278–11286,
  \doi{10.1609/aaai.v34i07.6788},
  \urlprefix\url{http://dx.doi.org/10.1609/aaai.v34i07.6788}

\bibitem[{Kim et~al(2023)Kim, Chae, Jang, and Yoon}]{kim2023event}
Kim T, Chae Y, Jang HK, Yoon KJ (2023) Event-based video frame interpolation
  with cross-modal asymmetric bidirectional motion fields. In: Proceedings of
  the IEEE/CVF Conference on Computer Vision and Pattern Recognition, pp
  18032--18042

\bibitem[{Kingma and Ba(2014)}]{kingma2014adam}
Kingma DP, Ba J (2014) Adam: A method for stochastic optimization. arXiv
  preprint arXiv:14126980

\bibitem[{Lai et~al(2018)Lai, Huang, Ahuja, and Yang}]{lai2018fast}
Lai WS, Huang JB, Ahuja N, Yang MH (2018) Fast and accurate image
  super-resolution with deep laplacian pyramid networks. IEEE Transactions on
  Pattern Analysis and Machine Intelligence 41(11):2599--2613

\bibitem[{Lee et~al(2020)Lee, Jung, tom Dieck, and Chung}]{lee2020experiencing}
Lee H, Jung TH, tom Dieck MC, Chung N (2020) Experiencing immersive virtual
  reality in museums. Information \& Management 57(5):103229

\bibitem[{Liang et~al(2021)Liang, Cao, Sun, Zhang, Van~Gool, and
  Timofte}]{Liang2021SwinIR}
Liang J, Cao J, Sun G, Zhang K, Van~Gool L, Timofte R (2021) Swinir: Image
  restoration using swin transformer. In: 2021 IEEE/CVF International
  Conference on Computer Vision Workshops (ICCVW),
  \doi{10.1109/iccvw54120.2021.00210},
  \urlprefix\url{http://dx.doi.org/10.1109/iccvw54120.2021.00210}

\bibitem[{Liang et~al(2022)Liang, Fan, Xiang, Ranjan, Ilg, Green, Cao, Zhang,
  Timofte, and Gool}]{Liang2022RSST}
Liang J, Fan Y, Xiang X, Ranjan R, Ilg E, Green S, Cao J, Zhang K, Timofte R,
  Gool L (2022) Recurrent video restoration transformer with guided deformable
  attention. Advances in Neural Information Processing Systems

\bibitem[{Liu et~al(2018)Liu, Wang, Fan, Liu, Wang, Chang, Wang, and
  Huang}]{liu2018learning}
Liu D, Wang Z, Fan Y, Liu X, Wang Z, Chang S, Wang X, Huang TS (2018) Learning
  temporal dynamics for video super-resolution: A deep learning approach. IEEE
  Transactions on Image Processing 27(7):3432--3445

\bibitem[{Liu et~al(2021)Liu, Lin, Cao, Hu, Wei, Zhang, Lin, and
  Guo}]{liu2021swin}
Liu Z, Lin Y, Cao Y, Hu H, Wei Y, Zhang Z, Lin S, Guo B (2021) Swin
  transformer: Hierarchical vision transformer using shifted windows. In:
  Proceedings of the IEEE/CVF International Conference on Computer Vision, pp
  10012--10022

\bibitem[{Liu et~al(2022{\natexlab{a}})Liu, Ning, Cao, Wei, Zhang, Lin, and
  Hu}]{Liu2022VSwin}
Liu Z, Ning J, Cao Y, Wei Y, Zhang Z, Lin S, Hu H (2022{\natexlab{a}}) Video
  swin transformer. In: 2022 IEEE/CVF Conference on Computer Vision and Pattern
  Recognition (CVPR), \doi{10.1109/cvpr52688.2022.00320},
  \urlprefix\url{http://dx.doi.org/10.1109/cvpr52688.2022.00320}

\bibitem[{Liu et~al(2022{\natexlab{b}})Liu, Ning, Cao, Wei, Zhang, Lin, and
  Hu}]{liu2022video}
Liu Z, Ning J, Cao Y, Wei Y, Zhang Z, Lin S, Hu H (2022{\natexlab{b}}) Video
  swin transformer. In: Proceedings of the IEEE/CVF Conference on Computer
  Vision and Pattern Recognition, pp 3202--3211

\bibitem[{Lu et~al(2023)Lu, Wang, Liu, Wang, and Wang}]{lu2023learning}
Lu Y, Wang Z, Liu M, Wang H, Wang L (2023) Learning spatial-temporal implicit
  neural representations for event-guided video super-resolution. In:
  Proceedings of the IEEE/CVF Conference on Computer Vision and Pattern
  Recognition, pp 1557--1567

\bibitem[{Lu et~al(2025{\natexlab{a}})Lu, Qian, Rao, Xiao, Chen, and
  Xiong}]{lu2025rgb}
Lu Y, Qian Y, Rao Z, Xiao J, Chen L, Xiong H (2025{\natexlab{a}}) Rgb-event
  isp: The dataset and benchmark. arXiv preprint arXiv:250119129

\bibitem[{Lu et~al(2025{\natexlab{b}})Lu, Xu, Li, Wang, Cui, Yao, and
  Xiong}]{lu2025events}
Lu Y, Xu X, Li P, Wang Y, Cui Y, Yao H, Xiong H (2025{\natexlab{b}}) From
  events to enhancement: A survey on event-based imaging technologies. arXiv
  preprint arXiv:250505488

\bibitem[{Meyer et~al(2018)Meyer, Djelouah, McWilliams, Sorkine-Hornung, Gross,
  and Schroers}]{meyer2018phasenet}
Meyer S, Djelouah A, McWilliams B, Sorkine-Hornung A, Gross M, Schroers C
  (2018) Phasenet for video frame interpolation. In: Proceedings of the IEEE
  Conference on Computer Vision and Pattern Recognition, pp 498--507

\bibitem[{Micikevicius et~al(2017)Micikevicius, Narang, Alben, Diamos, Elsen,
  Garcia, Ginsburg, Houston, Kuchaiev, Venkatesh et~al}]{micikevicius2017mixed}
Micikevicius P, Narang S, Alben J, Diamos G, Elsen E, Garcia D, Ginsburg B,
  Houston M, Kuchaiev O, Venkatesh G, et~al (2017) Mixed precision training.
  arXiv preprint arXiv:171003740

\bibitem[{Nah et~al(2017)Nah, Kim, and Lee}]{Nah_Kim_Lee_2017}
Nah S, Kim TH, Lee KM (2017) Deep multi-scale convolutional neural network for
  dynamic scene deblurring. In: 2017 IEEE Conference on Computer Vision and
  Pattern Recognition (CVPR), \doi{10.1109/cvpr.2017.35},
  \urlprefix\url{http://dx.doi.org/10.1109/cvpr.2017.35}

\bibitem[{Nah et~al(2019)Nah, Son, and Lee}]{nah2019recurrent}
Nah S, Son S, Lee KM (2019) Recurrent neural networks with intra-frame
  iterations for video deblurring. In: Proceedings of the IEEE/CVF Conference
  on Computer Vision and Pattern Recognition, pp 8102--8111

\bibitem[{Niklaus and Liu(2020)}]{niklaus2020softmax}
Niklaus S, Liu F (2020) Softmax splatting for video frame interpolation. In:
  Proceedings of the IEEE/CVF Conference on Computer Vision and Pattern
  Recognition, pp 5437--5446

\bibitem[{Niklaus et~al(2017)Niklaus, Mai, and Liu}]{niklaus2017video}
Niklaus S, Mai L, Liu F (2017) Video frame interpolation via adaptive
  convolution. In: Proceedings of the IEEE Conference on Computer Vision and
  Pattern Recognition, pp 670--679

\bibitem[{Paikin et~al(2021)Paikin, Ater, Shaul, and
  Soloveichik}]{paikin2021efi}
Paikin G, Ater Y, Shaul R, Soloveichik E (2021) Efi-net: Video frame
  interpolation from fusion of events and frames. In: Proceedings of the
  IEEE/CVF Conference on Computer Vision and Pattern Recognition, pp 1291--1301

\bibitem[{Pan et~al(2019)Pan, Scheerlinck, Yu, Hartley, Liu, and
  Dai}]{pan2019bringing}
Pan L, Scheerlinck C, Yu X, Hartley R, Liu M, Dai Y (2019) Bringing a blurry
  frame alive at high frame-rate with an event camera. In: Proceedings of the
  IEEE/CVF Conference on Computer Vision and Pattern Recognition, pp 6820--6829

\bibitem[{Park et~al(2021)Park, Lee, and Kim}]{park2021asymmetric}
Park J, Lee C, Kim CS (2021) Asymmetric bilateral motion estimation for video
  frame interpolation. In: Proceedings of the IEEE/CVF International Conference
  on Computer Vision, pp 14539--14548

\bibitem[{Parker(2010)}]{parker2010algorithms}
Parker JR (2010) Algorithms for image processing and computer vision. John
  Wiley \& Sons

\bibitem[{Paszke et~al(2019)Paszke, Gross, Massa, Lerer, Bradbury, Chanan,
  Killeen, Lin, Gimelshein, Antiga et~al}]{paszke2019pytorch}
Paszke A, Gross S, Massa F, Lerer A, Bradbury J, Chanan G, Killeen T, Lin Z,
  Gimelshein N, Antiga L, et~al (2019) Pytorch: An imperative style,
  high-performance deep learning library. Advances in Neural Information
  Processing Systems 32

\bibitem[{Ramasinghe and Lucey(2023)}]{ramasinghe2023learnable}
Ramasinghe S, Lucey S (2023) A learnable radial basis positional embedding for
  coordinate-mlps. In: Proceedings of the AAAI Conference on Artificial
  Intelligence, vol~37, pp 2137--2145

\bibitem[{Scheerlinck et~al(2019{\natexlab{a}})Scheerlinck, Rebecq, Stoffregen,
  Barnes, Mahony, and
  Scaramuzza}]{Scheerlinck_Rebecq_Stoffregen_Barnes_Mahony_Scaramuzza_2019}
Scheerlinck C, Rebecq H, Stoffregen T, Barnes N, Mahony R, Scaramuzza D
  (2019{\natexlab{a}}) Ced: Color event camera dataset. In: 2019 IEEE/CVF
  Conference on Computer Vision and Pattern Recognition Workshops (CVPRW),
  \doi{10.1109/cvprw.2019.00215},
  \urlprefix\url{http://dx.doi.org/10.1109/cvprw.2019.00215}

\bibitem[{Scheerlinck et~al(2019{\natexlab{b}})Scheerlinck, Rebecq, Stoffregen,
  Barnes, Mahony, and Scaramuzza}]{scheerlinck2019ced}
Scheerlinck C, Rebecq H, Stoffregen T, Barnes N, Mahony R, Scaramuzza D
  (2019{\natexlab{b}}) Ced: Color event camera dataset. In: Proceedings of the
  IEEE/CVF Conference on Computer Vision and Pattern Recognition Workshops, pp
  0--0

\bibitem[{Sironi et~al(2018)Sironi, Brambilla, Bourdis, Lagorce, and
  Benosman}]{Sironi2018TimeSurface}
Sironi A, Brambilla M, Bourdis N, Lagorce X, Benosman R (2018) Hats: Histograms
  of averaged time surfaces for robust event-based object classification. In:
  2018 IEEE/CVF Conference on Computer Vision and Pattern Recognition,
  \doi{10.1109/cvpr.2018.00186},
  \urlprefix\url{http://dx.doi.org/10.1109/cvpr.2018.00186}

\bibitem[{Song et~al(2022)Song, Huang, and Bajaj}]{song2022cir}
Song C, Huang Q, Bajaj C (2022) E-cir: Event-enhanced continuous intensity
  recovery. In: Proceedings of the IEEE/CVF Conference on Computer Vision and
  Pattern Recognition, pp 7803--7812

\bibitem[{Song et~al(2023)Song, Bajaj, and Huang}]{song2023deblursr}
Song C, Bajaj C, Huang Q (2023) Deblursr: Event-based motion deblurring under
  the spiking representation. arXiv preprint arXiv:230308977

\bibitem[{Su et~al(2017)Su, Delbracio, Wang, Sapiro, Heidrich, and
  Wang}]{Su_Delbracio_Wang_Sapiro_Heidrich_Wang_2017}
Su S, Delbracio M, Wang J, Sapiro G, Heidrich W, Wang O (2017) Deep video
  deblurring for hand-held cameras. In: 2017 IEEE Conference on Computer Vision
  and Pattern Recognition (CVPR), \doi{10.1109/cvpr.2017.33},
  \urlprefix\url{http://dx.doi.org/10.1109/cvpr.2017.33}

\bibitem[{Sun et~al(2022)Sun, Sakaridis, Liang, Jiang, Yang, Sun, Ye, Wang, and
  Gool}]{sun2022event}
Sun L, Sakaridis C, Liang J, Jiang Q, Yang K, Sun P, Ye Y, Wang K, Gool LV
  (2022) Event-based fusion for motion deblurring with cross-modal attention.
  In: European Conference on Computer Vision, Springer, pp 412--428

\bibitem[{Suzuki and Ikehara(2020)}]{suzuki2020residual}
Suzuki K, Ikehara M (2020) Residual learning of video frame interpolation using
  convolutional lstm. IEEE Access 8:134185--134193

\bibitem[{Tian et~al(2020)Tian, Zhang, Fu, and Xu}]{tian2020tdan}
Tian Y, Zhang Y, Fu Y, Xu C (2020) Tdan: Temporally-deformable alignment
  network for video super-resolution. In: Proceedings of the IEEE/CVF
  Conference on Computer Vision and Pattern Recognition, pp 3360--3369

\bibitem[{Tulyakov et~al(2021)Tulyakov, Gehrig, Georgoulis, Erbach, Gehrig, Li,
  and Scaramuzza}]{tulyakov2021time}
Tulyakov S, Gehrig D, Georgoulis S, Erbach J, Gehrig M, Li Y, Scaramuzza D
  (2021) Time lens: Event-based video frame interpolation. In: Proceedings of
  the IEEE/CVF Conference on Computer Vision and Pattern Recognition, pp
  16155--16164

\bibitem[{Tulyakov et~al(2022)Tulyakov, Bochicchio, Gehrig, Georgoulis, Li, and
  Scaramuzza}]{tulyakov2022time}
Tulyakov S, Bochicchio A, Gehrig D, Georgoulis S, Li Y, Scaramuzza D (2022)
  Time lens++: Event-based frame interpolation with parametric non-linear flow
  and multi-scale fusion. In: Proceedings of the IEEE/CVF Conference on
  Computer Vision and Pattern Recognition, pp 17755--17764

\bibitem[{Wang et~al(2020)Wang, Guo, Liu, Lin, Deng, and An}]{wang2020deep}
Wang L, Guo Y, Liu L, Lin Z, Deng X, An W (2020) Deep video super-resolution
  using hr optical flow estimation. IEEE Transactions on Image Processing
  29:4323--4336

\bibitem[{Wang et~al(2021)Wang, Zhang, Yuan, and Wang}]{wang2021unsupervised}
Wang W, Zhang H, Yuan Z, Wang C (2021) Unsupervised real-world
  super-resolution: A domain adaptation perspective. In: Proceedings of the
  IEEE/CVF International Conference on Computer Vision, pp 4318--4327

\bibitem[{Wang et~al(2019)Wang, Chan, Yu, Dong, and Change~Loy}]{wang2019edvr}
Wang X, Chan KC, Yu K, Dong C, Change~Loy C (2019) Edvr: Video restoration with
  enhanced deformable convolutional networks. In: Proceedings of the IEEE/CVF
  Conference on Computer Vision and Pattern Recognition workshops, pp 0--0

\bibitem[{Wang et~al(2004)Wang, Bovik, Sheikh, and Simoncelli}]{wang2004image}
Wang Z, Bovik AC, Sheikh HR, Simoncelli EP (2004) Image quality assessment:
  from error visibility to structural similarity. IEEE Transactions on Image
  Processing 13(4):600--612

\bibitem[{Wang et~al(2025)Wang, Hamann, Chaney, Jiang, Gallego, and
  Daniilidis}]{wang2025event}
Wang Z, Hamann F, Chaney K, Jiang W, Gallego G, Daniilidis K (2025) Event-based
  continuous color video decompression from single frames. In: Proceedings of
  the Computer Vision and Pattern Recognition Conference, pp 4968--4978

\bibitem[{Wei et~al(2025)Wei, Li, Tang, Zhao, and Bai}]{wei2025evenhancer}
Wei S, Li F, Tang S, Zhao Y, Bai H (2025) Evenhancer: Empowering effectiveness,
  efficiency and generalizability for continuous space-time video
  super-resolution with events. In: Proceedings of the Computer Vision and
  Pattern Recognition Conference, pp 17755--17766

\bibitem[{Xiang et~al(2020)Xiang, Tian, Zhang, Fu, Allebach, and
  Xu}]{xiang2020zooming}
Xiang X, Tian Y, Zhang Y, Fu Y, Allebach JP, Xu C (2020) Zooming slow-mo: Fast
  and accurate one-stage space-time video super-resolution. In: Proceedings of
  the IEEE/CVF Conference on Computer Vision and Pattern Recognition, pp
  3370--3379

\bibitem[{Xu et~al(2021)Xu, Xu, Li, Wang, Sun, and Cheng}]{xu2021temporal}
Xu G, Xu J, Li Z, Wang L, Sun X, Cheng MM (2021) Temporal modulation network
  for controllable space-time video super-resolution. In: 2021 IEEE/CVF
  Conference on Computer Vision and Pattern Recognition (CVPR),
  \doi{10.1109/cvpr46437.2021.00632},
  \urlprefix\url{http://dx.doi.org/10.1109/cvpr46437.2021.00632}

\bibitem[{Xu et~al(2019)Xu, Siyao, Sun, Yin, and Yang}]{xu2019quadratic}
Xu X, Siyao L, Sun W, Yin Q, Yang MH (2019) Quadratic video interpolation.
  Advances in Neural Information Processing Systems 32

\bibitem[{Xue et~al(2019)Xue, Chen, Wu, Wei, and Freeman}]{xue2019video}
Xue T, Chen B, Wu J, Wei D, Freeman WT (2019) Video enhancement with
  task-oriented flow. International Journal of Computer Vision 127:1106--1125

\bibitem[{Yan et~al(2025)Yan, Lu, Chen, Ma, Tang, Zheng, and
  Pan}]{yan2025evstvsr}
Yan H, Lu Z, Chen Z, Ma D, Tang H, Zheng Q, Pan G (2025) Evstvsr: Event guided
  space-time video super-resolution. In: Proceedings of the AAAI Conference on
  Artificial Intelligence, vol~39, pp 9085--9093

\bibitem[{Yang et~al(2021)Yang, Xiang, Zeng, and Zhang}]{yang2021real}
Yang X, Xiang W, Zeng H, Zhang L (2021) Real-world video super-resolution: A
  benchmark dataset and a decomposition based learning scheme. In: Proceedings
  of the IEEE/CVF International Conference on Computer Vision, pp 4781--4790

\bibitem[{Yue et~al(2022)Yue, Zhang, and Yang}]{yue2022real}
Yue H, Zhang Z, Yang J (2022) Real-rawvsr: Real-world raw video
  super-resolution with a benchmark dataset. arXiv preprint arXiv:220912475

\bibitem[{Zhang(2020)}]{zhang2020and}
Zhang C (2020) The why, what, and how of immersive experience. Ieee Access
  8:90878--90888

\bibitem[{Zhang et~al(2018)Zhang, Isola, Efros, Shechtman, and
  Wang}]{zhang2018unreasonable}
Zhang R, Isola P, Efros AA, Shechtman E, Wang O (2018) The unreasonable
  effectiveness of deep features as a perceptual metric. In: Proceedings of the
  IEEE Conference on Computer Vision and Pattern Recognition, pp 586--595

\bibitem[{Zhang and Yu(2022)}]{zhang2022unifying}
Zhang X, Yu L (2022) Unifying motion deblurring and frame interpolation with
  events. In: Proceedings of the IEEE/CVF Conference on Computer Vision and
  Pattern Recognition, pp 17765--17774

\bibitem[{Zhang et~al(2020)Zhang, Wang, and Tao}]{zhang2020video}
Zhang Y, Wang C, Tao D (2020) Video frame interpolation without temporal
  priors. Advances in Neural Information Processing Systems 33:13308--13318

\bibitem[{Zhang et~al(2022)Zhang, Wang, Zhu, and Chen}]{zhang2022optical}
Zhang Y, Wang H, Zhu H, Chen Z (2022) Optical flow reusing for high-efficiency
  space-time video super resolution. IEEE Transactions on Circuits and Systems
  for Video Technology 33(5):2116--2128

\bibitem[{Zhao et~al(2019)Zhao, Zheng, Xu, and Wu}]{zhao2019object}
Zhao ZQ, Zheng P, Xu St, Wu X (2019) Object detection with deep learning: A
  review. IEEE transactions on neural networks and learning systems
  30(11):3212--3232

\bibitem[{Zheng et~al(2023)Zheng, Liu, Lu, Hua, Pan, Zhang, Tao, and
  Wang}]{zheng2023deep}
Zheng X, Liu Y, Lu Y, Hua T, Pan T, Zhang W, Tao D, Wang L (2023) Deep learning
  for event-based vision: A comprehensive survey and benchmarks. arXiv preprint
  arXiv:230208890

\bibitem[{Zou et~al(2023)Zou, Chen, Shi, Guo, and Ye}]{zou2023object}
Zou Z, Chen K, Shi Z, Guo Y, Ye J (2023) Object detection in 20 years: A
  survey. Proceedings of the IEEE

\end{thebibliography}
}

\clearpage
\appendix
\clearpage

\section*{Supplementary Materials Preview:}
Additional videos have been included in the supplementary materials to provide a more comprehensive demonstration of our method's visual results. 
Below, we enumerate these videos and briefly describe their key features.
We then present more visualizations to demonstrate the generalization of our method on real data.

\begin{itemize}
    \item \texttt{1-Adobe240:} This video contains the following five clips.
    \begin{itemize}
        \item \texttt{IMG\_0013-7skip4xsr-Cyclist:} In this video, the camera and the \textbf{people in the background} are in motion, creating a complex scene. Our method successfully recovers the \textbf{locally moving bicycle}, demonstrating exceptional video frame interpolation and super-resolution capabilities.
        \item \texttt{IMG\_0037-7skip4xsr-\\TrafficIntersectionManyCars:} The video demonstrates a \textbf{camera with slight movement}, capturing a \textbf{busy intersection bustling with vehicles}. Our method is capable of accurately recovering vehicles in motion within the scene, including the intricate details of rotating tires.
        \item \texttt{IMG\_0037a-7skip-\\MovingForegroundAndBackground:} The video includes both \textbf{distant and close-up elements}. In the close-up scenes, the comparative methods resulted in significant deformations and distortions.
        \item \texttt{IMG\_0045-7skip-\\PortraitSculpture:} This video demonstrates the effects under \textbf{significant camera movement}. When the camera moves rapidly, frame-based methods tend to underperform.
        \item \texttt{IMG\_0175-7skip4xsr-\\LawnAndCar:} The same scene occurs when the \textbf{camera moves violently}.
        \item \texttt{IMG\_0175-7skip4xsr-\\TreeComplexTexture:} This video captures leaves, demonstrating that methods based on optical flow tend to fail in the presence of complex textures.
    \end{itemize} 
    \item \texttt{2-TimeLensPP-Ours-1:} This video shows the performance of our method on real-world data sets and the visualization of features. Demonstrates that we effectively capture local motion.
    \item \texttt{3-Our-vs-Timelens:} This video shows the results of comparing our method with Timelens.
\end{itemize}

\begin{figure*}
    \centering
    \includegraphics[width=0.9\linewidth]{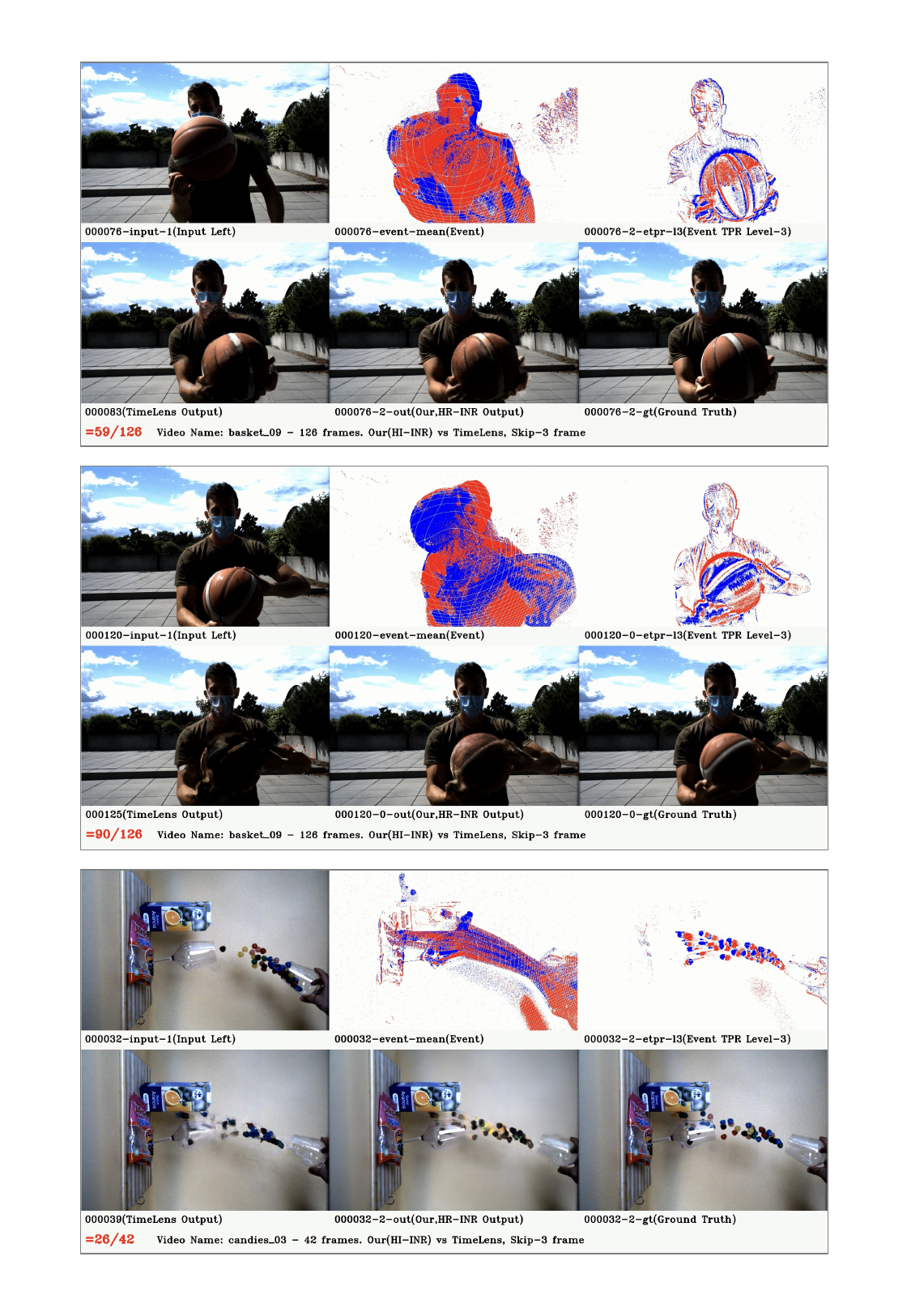}
    \caption{More visualization results on real-world data set~\citeps{tulyakov2021time}.}
\end{figure*}
\begin{figure*}
    \centering
    \includegraphics[width=0.9\linewidth]{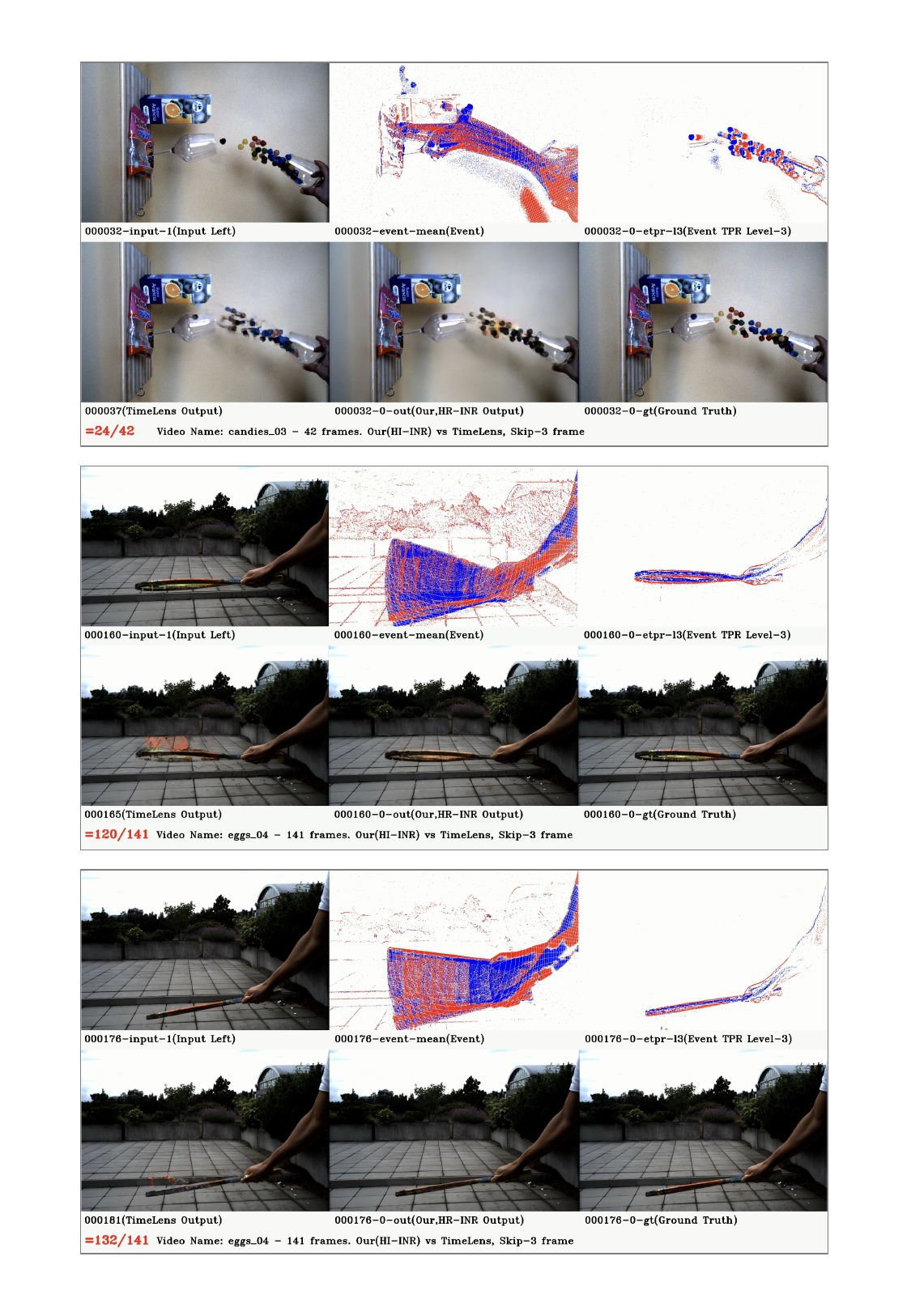}
    \caption{More visualization results on real-world data set~\citeps{tulyakov2021time}.}
\end{figure*}
\begin{figure*}
    \centering
    \includegraphics[width=0.9\linewidth]{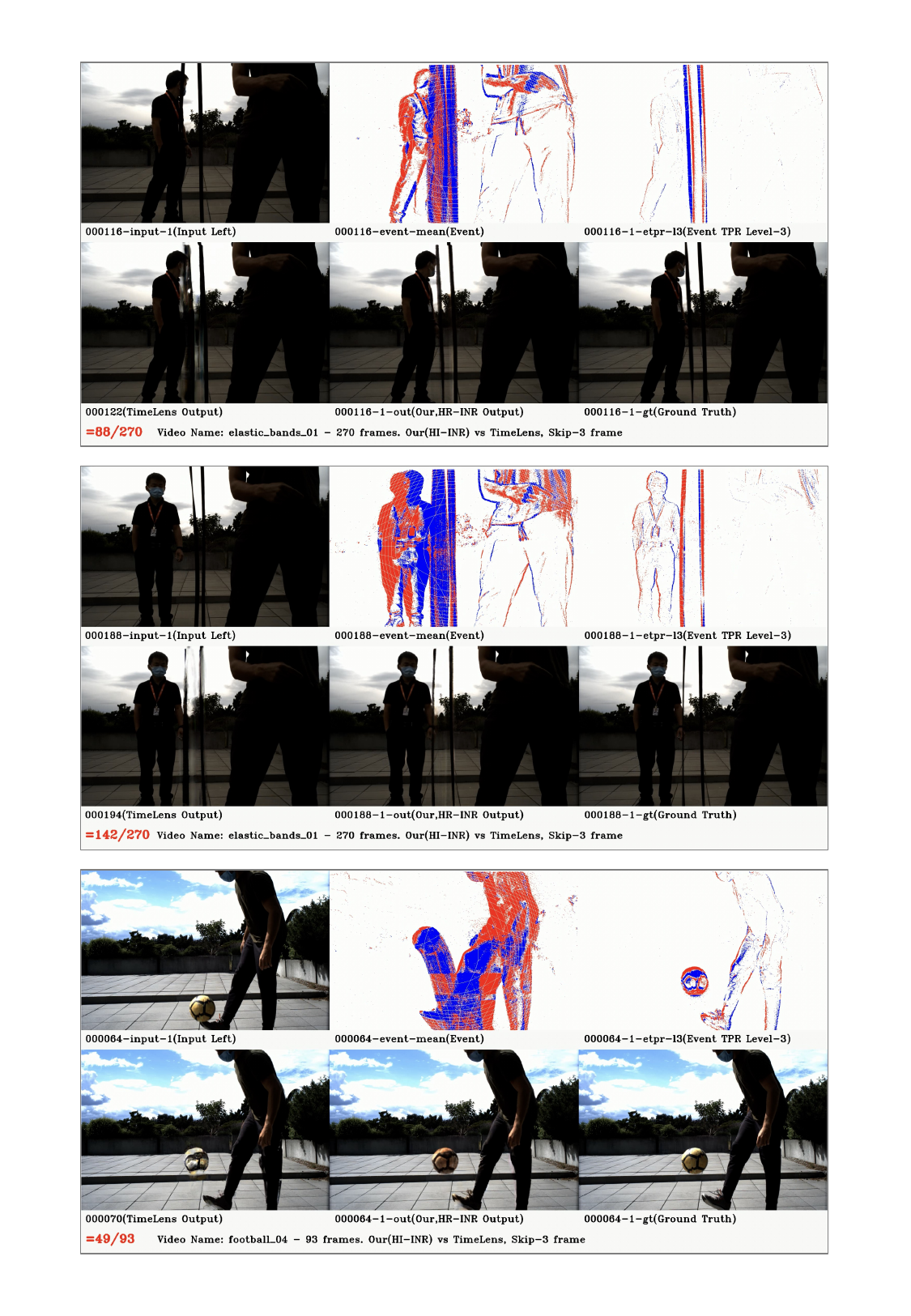}
    \caption{More visualization results on real-world data set~\citeps{tulyakov2021time}.}
\end{figure*}
\begin{figure*}
    \centering
    \includegraphics[width=0.9\linewidth]{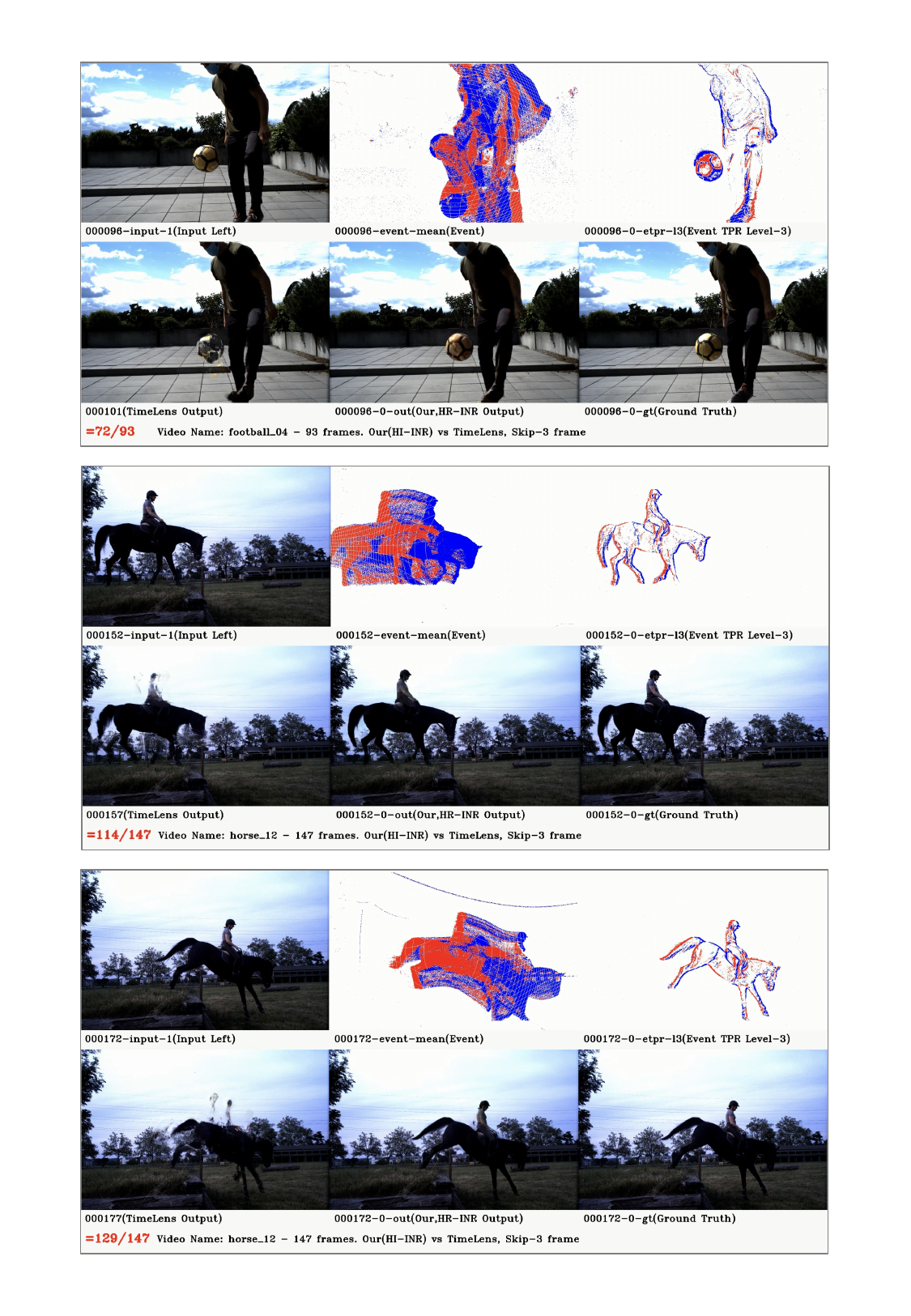}
    \caption{More visualization results on real-world data set~\citeps{tulyakov2021time}.}
\end{figure*}
\begin{figure*}
    \centering
    \includegraphics[width=0.9\linewidth]{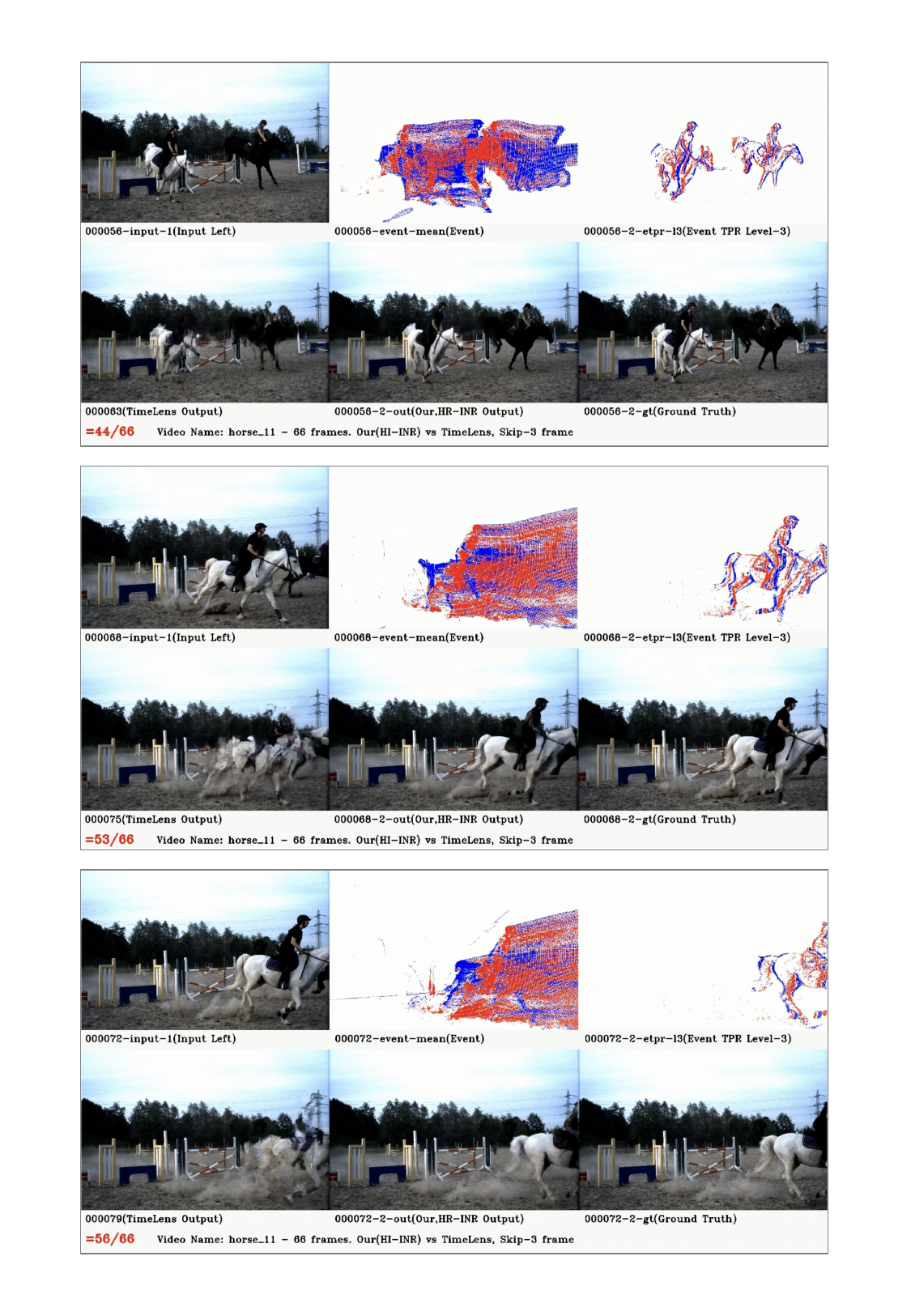}
    \caption{More visualization results on real-world data set~\citeps{tulyakov2021time}.}
\end{figure*}
\begin{figure*}
    \centering
    \includegraphics[width=0.9\linewidth]{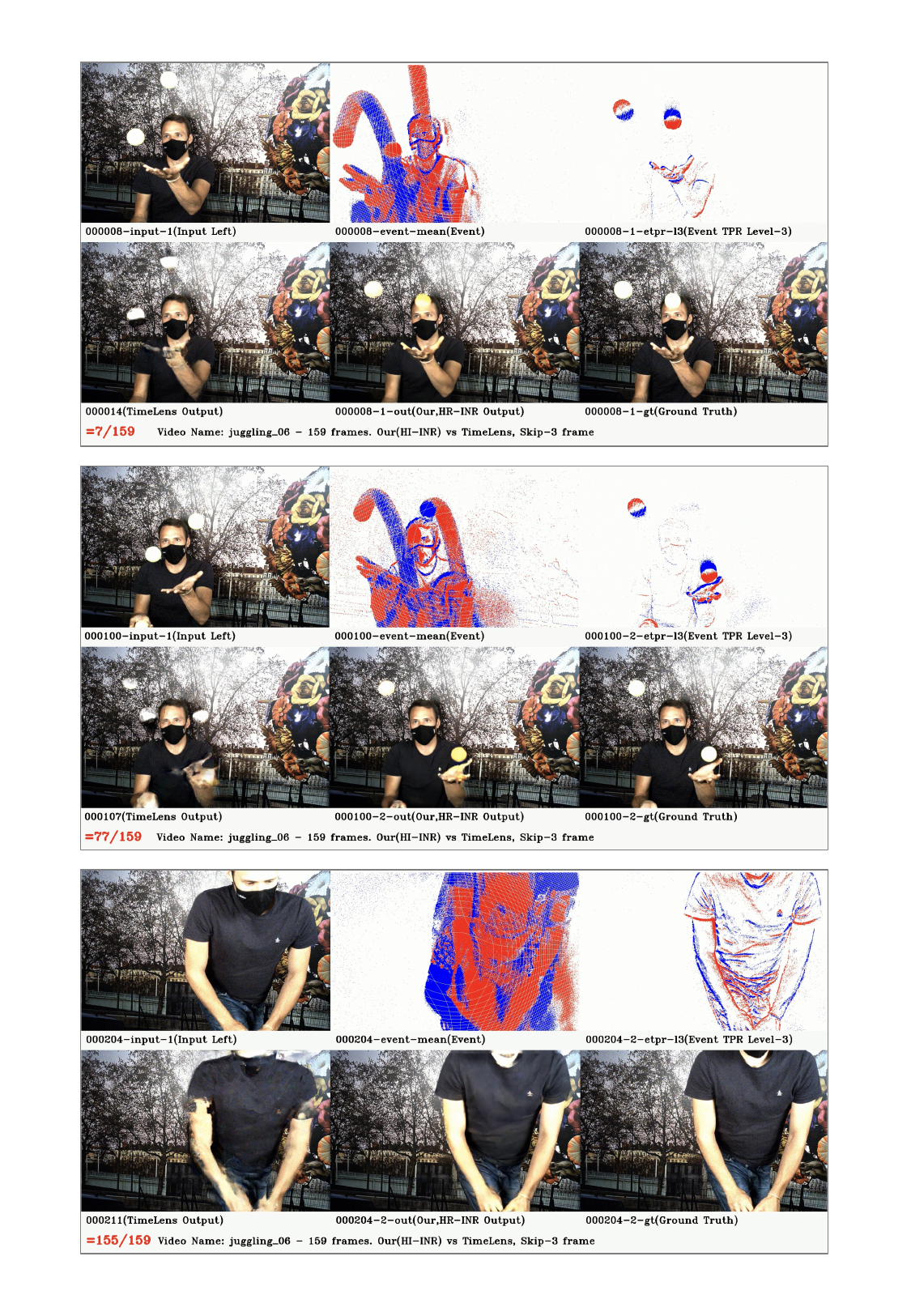}
    \caption{More visualization results on real-world data set~\citeps{tulyakov2021time}.}
\end{figure*}
\begin{figure*}
    \centering
    \includegraphics[width=0.9\linewidth]{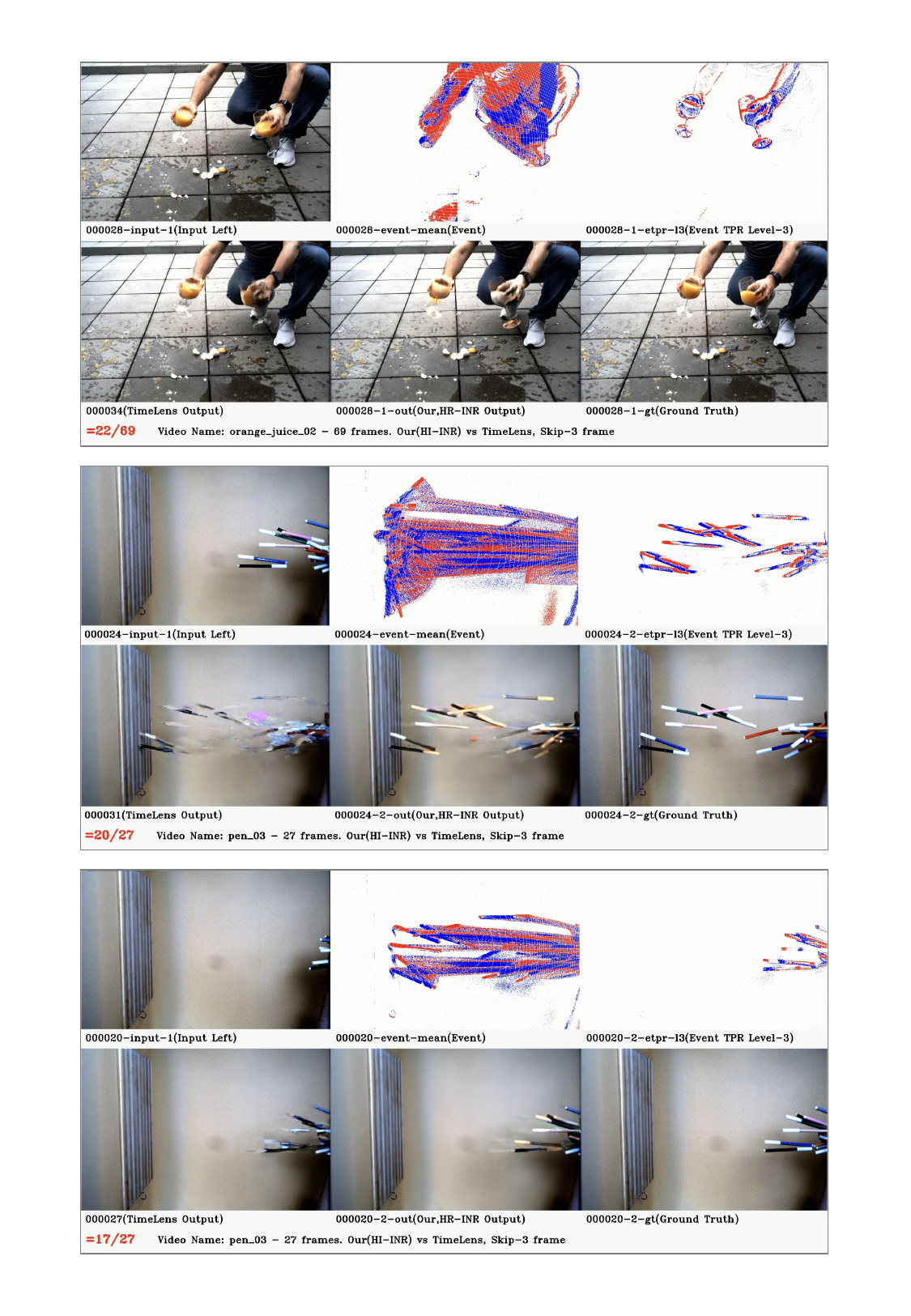}
    \caption{More visualization results on real-world data set~\citeps{tulyakov2021time}.}
\end{figure*}

\end{document}